\documentclass[conference]{IEEEtran}
\IEEEoverridecommandlockouts
\usepackage{cite}
\usepackage{stackengine}
\usepackage{subfig}
\usepackage{amsmath,amssymb,amsfonts}
\usepackage{algorithmic,algorithm}
\usepackage{graphicx}
\usepackage{textcomp}
\usepackage{xcolor}
\usepackage{hyperref}
\usepackage{balance}
\usepackage{caption}

\def\BibTeX{{\rm B\kern-.05em{\sc i\kern-.025em b}\kern-.08em
    T\kern-.1667em\lower.7ex\hbox{E}\kern-.125emX}}
    
\begin{document}

\title{ChronoGAN: Supervised and Embedded Generative Adversarial Networks for Time Series Generation
}

\author{
    \IEEEauthorblockN{MohammadReza EskandariNasab, Shah Muhammad Hamdi, Soukaina Filali Boubrahimi}
    \IEEEauthorblockA{
        \textit{Department of Computer Science}, \\
        \textit{Utah State University}, \\
        Logan, UT 84322, USA \\
        \{reza.eskandarinasab, s.hamdi, soukaina.boubrahimi\}@usu.edu
    }
}

\maketitle

\begin{abstract}
Generating time series data using Generative Adversarial Networks (GANs) presents several prevalent challenges, such as slow convergence, information loss in embedding spaces, instability, and performance variability depending on the series length. To tackle these obstacles, we introduce a robust framework aimed at addressing and mitigating these issues effectively. This advanced framework integrates the benefits of an Autoencoder-generated embedding space with the adversarial training dynamics of GANs. This framework benefits from a time series-based loss function and oversight from a supervisory network, both of which capture the stepwise conditional distributions of the data effectively. The generator functions within the latent space, while the discriminator offers essential feedback based on the feature space. Moreover, we introduce an early generation algorithm and an improved neural network architecture to enhance stability and ensure effective generalization across both short and long time series. Through joint training, our framework consistently outperforms existing benchmarks, generating high-quality time series data across a range of real and synthetic datasets with diverse characteristics.
\end{abstract}

\begin{IEEEkeywords}
Time Series Generation, Generative Adversarial Networks, Autoencoders, Data Augmentation
\end{IEEEkeywords}

\section{Introduction}
Fields such as biomedical signal processing \cite{eskandarinasab2024grucnn} and solar flare prediction \cite{hamdiflare, velanki2024} often face data shortages due to complex and noisy data environments, scarcity of events, and privacy concerns \cite{privacy}, all of which complicate accurate model training and evaluation. Developing methods that leverage Generative Adversarial Networks (GANs) \cite{goodfellow2014generative} to produce realistic synthetic data can foster scientific progress. By creating balanced datasets and mitigating data shortages, GANs can improve the performance of machine learning tasks \cite{ahmadzadeh2021train}.

Generative modeling of time series data poses unique challenges due to the temporal nature of the data. These models must not only capture the distribution of features at individual time points but also unravel the complex dynamics between these points over time. For instance, when managing multivariate sequential data represented as $x_{1:T} = (x_1, \ldots, x_T)$, an effective model should accurately determine the conditional distribution $p(x_t \mid x_{1:t-1})$, which dictates the temporal transitions. Without this capability, the generated data fails to capture the characteristics of the real dataset \cite{obahri}. This leads to misleading and inaccurate evaluations when used alongside real data for downstream machine learning tasks \cite{angryk2020multivariate}.

In the field of time series generation, a substantial body of research has focused on enhancing the temporal dynamics of autoregressive models for sequence forecasting. The primary aim is to reduce the propagation of sampling errors through various training-time adjustments, leading to more precise conditional distribution modeling \cite{NIPS2015_e995f98d, lamb2016professor, bahdanau2017actorcritic}. Autoregressive models decompose the sequence distribution into a chain of conditionals, \(\prod_{t} p(x_t \mid x_{1:t-1})\), which proves useful for forecasting due to their deterministic nature. However, they lack true generative capabilities, as generating new sequences from them does not require external input. In contrast, research applying GANs to sequential data often employs sequence-to-sequence neural network layers for both the generator and discriminator. This approach pursues a direct adversarial objective \cite{mogren2016crnngan, esteban2017realvalued, ramponi2019tcgan} to learn the probability distribution of the data and generate new samples by feeding random noise into the model. While straightforward, this adversarial goal focuses on modeling the joint distribution \(p(x_{1:T})\) \cite{peiyu} without considering the autoregressive structure. This may be inadequate, as aggregating standard GAN losses across vectors does not necessarily ensure the capture of stepwise dependencies in time series samples.

In this paper, we introduce a novel framework that significantly enhances stability, accuracy, and generalizability. Our approach, termed ChronoGAN, effectively integrates the two research streams into a robust and precise generative model specifically designed to preserve temporal dynamics through supervised GAN training. Additionally, it leverages latent space during training, ensuring more reliable convergence. Therefore, ChronoGAN offers a comprehensive method for generating realistic time-series data applicable across various fields. The key contributions of our study are:

\begin{enumerate}
    \item Generating data within the latent space using a generator, while utilizing a discriminator that operates in the feature space, offers significant advantages. This method not only provides more precise adversarial feedback to the generator but also delivers crucial adversarial feedback to the autoencoder, enhancing the overall performance of the model.
    \item The development of a novel time series-based loss function for the generator network, combined with a supervised loss, enhances the quality of the generated data by more effectively learning the temporal dynamics. Additionally, a new loss function is designed for the autoencoder to improve its reconstruction capabilities.
    \item The implementation of an early generation algorithm to stabilize the framework and ensure optimal results after each training session.
    \item The implementation of a novel GRU-LSTM architecture across the framework's five neural networks to enhance the generation of high-quality data for sequences of varying lengths, both short and long.
\end{enumerate}

We demonstrate the advantages of ChronoGAN by conducting a series of experiments on a variety of real-world and synthetic datasets. Our findings indicate that ChronoGAN consistently outperforms existing benchmarks, including TimeGAN \cite{NEURIPS2019_c9efe5f2}, in generating realistic time-series data.

\section{Related Work}
\label{sec:relatedwork}

Autoregressive recurrent networks trained using maximum likelihood methods are susceptible to significant prediction errors during multi-step sampling \cite{Williams1989}. This issue arises from the difference between closed-loop training (conditioned on actual data) and open-loop inference (based on prior predictions). Further, inspired by adversarial domain adaptation \cite{Pforcing}, Professor Forcing trains an additional discriminator to differentiate between autonomous and teacher-driven hidden states \cite{Tforcing}, helping to align training and sampling dynamics. However, although these methods share our aim of modeling stepwise transitions, they are deterministic and do not explicitly involve sampling from a learned distribution, which is crucial for our objective of synthetic data generation.

The foundational paper on GANs \cite{goodfellow2014generative} introduced a novel framework for generating synthetic data. The model consists of two neural networks (the generator and the discriminator) that are trained simultaneously in a zero-sum game setup. However, despite being capable of generating data by sampling from a learned distribution, they struggle to capture the stepwise dependencies inherent in time series data. The adversarial feedback from the discriminator alone is insufficient for the generator to effectively learn the patterns of sequences.

Several studies have adopted the GAN framework for use in time series analysis. The earliest, C-RNN-GAN \cite{mogren2016crnngan}, applied the GAN directly to sequential data with LSTM networks serving as both generator and discriminator. It generates data recurrently, starting with a noise vector and the data from the previous time step. RCGAN \cite{esteban2017realvalued} modified this by removing the reliance on previous outputs and incorporating additional inputs for conditioning \cite{mirza2014conditional}. However, these models depend solely on binary adversarial feedback for learning, which may not capture the temporal dynamics of time series data.

TimeGAN \cite{NEURIPS2019_c9efe5f2} presented a sophisticated method for generating time-series data, combining the versatility of unsupervised learning with the accuracy of supervised training. By optimizing an embedding space through both supervised and adversarial objectives, it aimed to closely mirror the dynamics of time series data. Despite its novel approach, TimeGAN encounters challenges with the quality of the generated data, primarily due to its reliance on adversarial training within the embedding space rather than the feature space. Furthermore, TimeGAN suffers from stability issues, yielding inconsistent outcomes across identical iteration counts and hyperparameter settings. It also faces difficulties in generating both short and long time series sequences.

The ChronoGAN framework is developed to enhance the efficacy and robustness of time series generation by accomplishing several critical objectives. First, it optimizes performance across both short and long sequences. Second, it enhances data reconstruction by the decoder and data generation by the generator through providing more accurate adversarial feedback to both the autoencoder and generator. Third, it facilitates the convergence of both the generator and autoencoder networks through the implementation of novel loss functions. Finally, it incorporates an early generation algorithm to achieve consistent optimal results under the same hyperparameters. Fig. \ref{fig:chronogan} illustrates the implementation of ChronoGAN.

\section{Proposed Model: ChronoGAN}
\label{sec:method}

\begin{figure}
  \centering
  \includegraphics[width=0.5\textwidth]{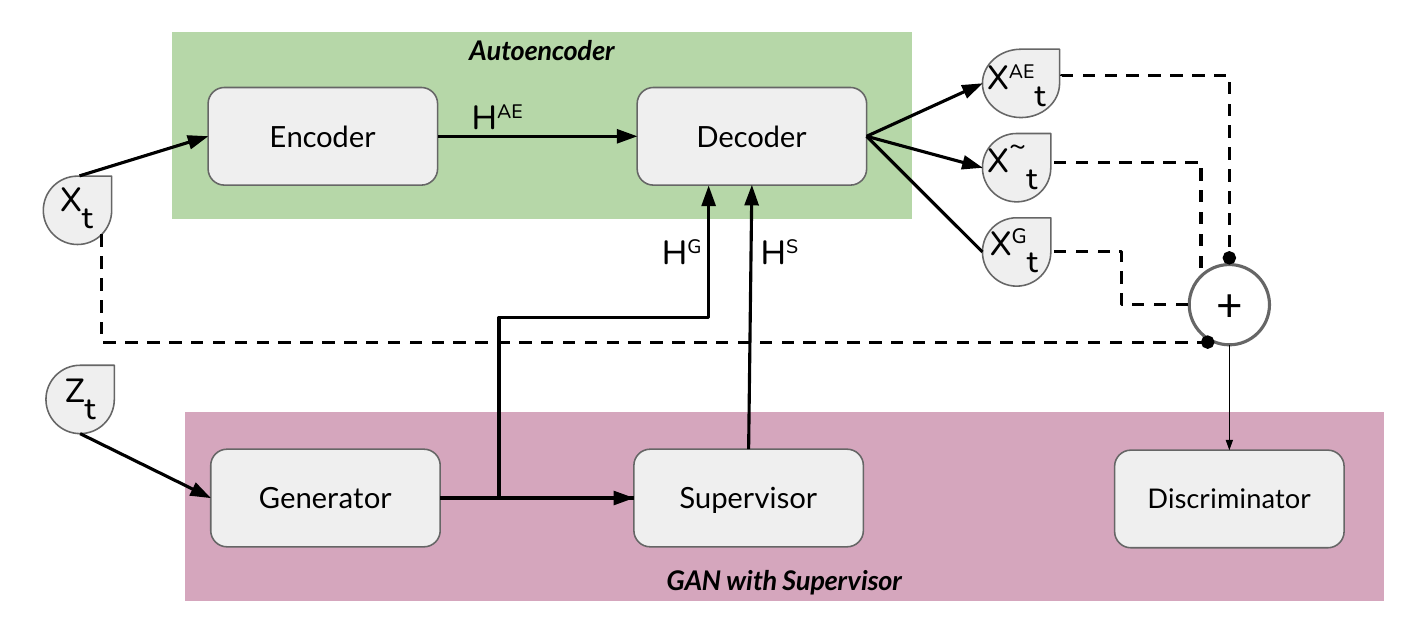}
  \caption{The figure illustrates the architecture of ChronoGAN for time series generation. ChronoGAN consists of five neural networks, each utilizing sequence-to-sequence GRU-LSTM layers. These networks are trained jointly to learn the probability distribution of real data and to capture the temporal dynamics inherent in the real samples.}
  \label{fig:chronogan}
\end{figure}

Based on Fig. \ref{fig:chronogan}, the framework includes five networks: an autoencoder (encoder and decoder), a generator, a supervisor, and a discriminator. The autoencoder's role is to facilitate training by generating compressed representations in the latent space, thereby reducing the likelihood of non-convergence within the GAN framework. The generator produces data in this lower-dimensional latent space, as opposed to the feature space. The supervisor network, integrated with a supervised loss function, is specifically designed to learn the temporal dynamics of the time series data. This is crucial, as sole reliance on the discriminator’s adversarial feedback may not sufficiently prompt the generator to capture the data’s stepwise conditional distributions. The discriminator network differentiates between fake and real data in the feature space, providing more accurate feedback to both the generator and autoencoder.

In Fig. \ref{fig:chronogan}, $H^{AE} = e(X)$ represents the encoding of the input data $X$ into a latent space $H^{AE}$ using the encoder function $e$. The reconstructed data $X^{AE} = r(H^{AE})$ is obtained by decoding $H^{AE}$ using the recovery function $r$, aiming to replicate the original input data as closely as possible. The generator function \( g \) transforms a random noise vector \( Z \) into synthetic latent data \( H^{G} = g(Z) \), which is then reconstructed into synthetic data \( X^{G} = r(H^{G}) \). The supervisor network $s$ processes $H^{G}$ to produce a supervised latent representation $H^{S} = s(H^{G})$, from which the final synthetic data $\tilde{X} = r(H^{S})$ is reconstructed. The discriminator $d$ evaluates the authenticity of the synthetic and real data by outputting $\tilde{y}$ for synthetic data and $y$ for real data.

\subsection{Adversarial Training}

In a joint training scheme involving a GAN network and an autoencoder, relying solely on reconstruction loss for the autoencoder results in noisy outputs, where the autoencoder's output fails to fully retain the input's characteristics \cite{noise}. Additionally, adversarial training within an embedding space leads to the generation of noisy data after decoding the generator’s output. The issue arises when the encoder's output (\(H^{AE}\)) is regarded as real data and the generator's output (\(H^{G}\)) as synthetic during the adversarial training process. This practice reduces the discriminator's ability to accurately differentiate between the attributes of real and synthetic data. A significant limitation is that the discriminator does not account for the error rate and data loss inherent in the autoencoder's performance. This oversight may compromise the efficacy of the discriminator, resulting in suboptimal performance in distinguishing between real and generated data attributes. Consequently, this leads to less precise feedback being provided to the generator network, potentially affecting the overall quality of the synthetic data. To address this, as shown in Fig. \ref{fig:chronogan}, discriminating in the feature space allows for defining real data as the dataset (\(X\)) and fake data as the decoding of the generator's output (\(X^{G}\)). This facilitates more accurate training for the discriminator, thus yielding improved feedback for the generator. Additionally, discrimination in the feature space provides valuable adversarial feedback to the autoencoder, enhancing its reconstruction capabilities in conjunction with conventional reconstruction loss. In the context of time series data, the feature space denotes the original dimensions, such as individual time points and their observed values. The latent or embedding space, achieved through an encoding process, represents the data in a lower-dimensional form, capturing its essential patterns and structures in a more compact and informative manner \cite{embed}.

Through a joint learning scheme, the autoencoder is initially trained using a combination of reconstruction loss and binary feedback from the discriminator, where real data is the dataset (\(X\)) and fake data is its reconstruction (\(X^{AE}\)). This approach enhances the autoencoder's precision in reconstructing outputs. In the subsequent phase, only the supervisor network is trained. The supervisor utilizes real data embeddings from the previous two time steps \( h_{1:t-2} \) generated by the embedding network to create the subsequent latent vector \( h_{t} \). Finally, all five networks are trained jointly. During this final phase, the same discriminator distinguishes between real data, denoted as the dataset ($X$), and the dataset reconstructions ($X^{AE}$), where the fake data comprises the generator's decoded outputs ($X^{G}$) and the supervisor’s decoded outputs ($\tilde{X}$). The generator undergoes training through this adversarial feedback $\mathcal{L}_U$, in addition to other feedback mechanisms including $\mathcal{L}_S$, $\mathcal{L}_V$, and $\mathcal{L}_{TS}$. This phase involves a shift in the characterization of fake and real data compared to the initial phase.

\subsection{Novel Loss Functions}

Based on the feedback from the discriminator, we introduce a new loss function for the autoencoder (\(\mathcal{L}_{AE}\)), which comprises both reconstruction loss (\(\mathcal{L}_{R}\)) and adversarial loss (\(\mathcal{L}_{U}\)). The proportion of reconstruction loss to adversarial loss decreases in the third phase of training compared to the first phase, where the primary purpose of the discriminator is to provide feedback for the generator rather than the autoencoder.

\begin{equation}
\mathcal{L}_{AE}
 = \mathcal{L}_{R} + \mathcal{L}_{U}
 \textbf{;} \quad 
\mathcal{L}_{R} = \mathbb{E}_{x_{1:T}\sim p} \left[ \sum_t \|\mathbf{x}_t - \mathbf{{x}^{AE}}_t\|_2 \right] 
\end{equation}

Where \( t \) denotes an individual time step, and \( T \) represents the total number of time steps within the series. In addition, $\mathbf{x}_t$ represents the real data at timestamp $t$, and $\mathbf{x}^{AE}_t$ denotes the output of the autoencoder corresponding to the real data $\mathbf{x}_t$ at the same timestamp.

\begin{equation}
\mathcal{L}_{U} = \mathbb{E}_{x_{1:T}\sim p} \left[ \sum_t \log y_t \right] + \mathbb{E}_{x_{1:T}\sim \tilde{p}} \left[ \sum_t \log (1 - \tilde{y}_t) \right]
\end{equation}

\begin{equation}
\tilde{y} = d(X^{AE}) \textbf{;} \quad y = d(X)
\end{equation}

Here, $p$ indicates the probability distribution of real data, and $\tilde{p}$ represents the probability distribution of synthetic data. Moreover, the discriminator \( d \) generates the output \( \tilde{y} \) when evaluating the autoencoder's output \( X^{AE} \) and produces the output \( y \) when assessing the real samples \( X \)

The sole reliance on the discriminator's binary adversarial feedback might not sufficiently drive the generator to capture the data's stepwise conditional distributions. To address this, ChronoGAN introduces an additional component, the supervisor, along with a novel loss mechanism denoted by \(\mathcal{L}_S\). ChronoGAN employs a closed-loop training mode, where the supervisor utilizes actual data embeddings from the previous two time steps \( h_{1:t-2} \) produced by the embedding network to generate the subsequent latent vector \( h_{t} \). This looped training involves the generator's loss \( \mathcal{L}_G \), which encompasses the adversarial loss \( \mathcal{L}_{U} \), the stepwise transition loss \( \mathcal{L}_S \), the distribution loss \(\mathcal{L}_V\), and our innovative time series loss \(\mathcal{L}_{TS}\). This structure ensures the generation of realistic sequences with accurate temporal transitions. The distribution loss \(\mathcal{L}_V\) leverages the mean absolute error (MAE) of the mean and variance between the real data \(X\) and the generated data \(\tilde{X}\). This approach effectively assists the generator in learning the real data distribution, enabling it to produce data across the entire distribution, which also serves as a key metric for evaluating GAN techniques.

\begin{equation}
\mathcal{L}_{G} = \mathcal{L}_{U} + \mathcal{L}_S + \mathcal{L}_V + \mathcal{L}_{TS}
 \textbf{;} \quad 
 \mathcal{L}_V = \mathcal{L}_{Mean} + \mathcal{L}_{Variance}
\end{equation}

Where $\mathcal{L}_{Mean}$ is the MAE of the mean between a batch of real and generated samples, and $\mathcal{L}_{Variance}$ is the MAE of the variance between the same batch of real and generated data.

\begin{equation}
\mathcal{L}_{\text{Mean}} = \mathbb{E}_{x_{1:T} \sim p} \left[ \sum_t \left| \frac{1}{N} \sum_{n=1}^N\mathbf{x}_{t_n} - \frac{1}{N} \sum_{n=1}^N  \mathbf{\tilde{x}_{t_n}} \right| \right]
\end{equation}

Where each sample is labeled by $n \in \{1, \ldots, N\}$ and the batch is represented as $\mathcal{B} = \{\mathbf{x}_{n,1:T_n}\}_{n=1}^{N}$.

\begin{align}
\mathcal{L}_{\text{Variance}} = & \, \mathbb{E}_{x_{1:T} \sim p} \left[ \sum_t \left| \frac{1}{N} \sum_{n=1}^N ( \mathbf{x}_{t_n} - \overline{\mathbf{x_t}})^2 \right. \right. \nonumber \\
& \left. \left. - \frac{1}{N} \sum_{n=1}^N(\mathbf{\tilde{x}}_{t_n} - \overline{\mathbf{\tilde{x_t}}})^2 \right| \right]
\end{align}

Where $\overline{\mathbf{x}}$ indicates the mean of $\mathbf{x}$, and $\overline{\mathbf{\tilde{x}}}$ represents the mean of $\tilde{\mathbf{x}}$ for a batch of data.

\begin{equation}
\mathcal{L}_S = \mathbb{E}_{x_{1:T}\sim p} \left[ \sum_t \left\| h^G_t - s(h^G_{t-2}) \right\|_2 \right]
\end{equation}

Where $s$ is the supervisor network, $h^G_t$ is the output of the generator at timestamp $t$, and $h^G_{t-2}$ is the output of the generator at timestamp $t-2$. This technique is more efficient than predicting timestamp $t$ using timestamp $t-1$.

In the third phase of training, referred to as joint training, $\tilde{y}$ represents the output of the discriminator $d$ for synthetic samples $X^G$ and $\tilde{X}$, while $y$ denotes the output of $d$ for real samples $X$ and $X^{AE}$.

\begin{equation}
\tilde{y} = d(X^{G}, \tilde{X}) \textbf{;} \quad {y} = d(X, X^{AE})
\end{equation}

Furthermore, we introduce a novel loss function for the generator called the time series loss, \(\mathcal{L}_{TS}\), which not only facilitates convergence but also enhances the quality of the generated data. This loss function is defined as the mean squared error (MSE) of the mean and standard deviation (std) of four key time series characteristics, including slope, skewness, weighted average, and median, between real and synthetic data. The aim is to boost the generator's convergence and its ability to learn the real data characteristics and distribution, as relying solely on the adversarial loss is insufficient for learning the characteristics of real time series data. The time series loss \(\mathcal{L}_{TS}\) is a novel contribution, comprising the slope loss (\(\mathcal{L}_{Slope}\)), weighted average loss (\(\mathcal{L}_{WeightedAvg}\)), skewness loss (\(\mathcal{L}_{Skewness}\)), and median loss (\(\mathcal{L}_{Median}\)).

\begin{equation}
\mathcal{L}_{TS} = \mathcal{L}_{Slope} + \mathcal{L}_{WeightedAvg} + \mathcal{L}_{Skewness} + \mathcal{L}_{Median}
\end{equation}

 The slope loss \(\mathcal{L}_{Slope}\) includes the MSE of the mean (\(\mathcal{L}_{S_{mean}}\)) and the MSE of the std (\(\mathcal{L}_{S_{std}}\)) between the slopes of real and generated samples.

\begin{equation}
\label{equ:ten}
\mathcal{L}_{Slope} = \mathcal{L}_{S_{mean}} + \mathcal{L}_{S_{std}}
\end{equation}

The slope is calculated using the provided formula,

\begin{equation}
\text{slope} = \frac{T \sum_{t=1}^T t x_t - \sum_{t=1}^T t \sum_{t=1}^T x_t}{T \sum_{t=1}^T t^2 - (\sum_{t=1}^T t)^2}
\end{equation}

In these equations, \(S\) is the slope of real samples, and \(\tilde{S}\) is the slope of generated samples.

\begin{equation}
\label{equ:twelve}
\mathcal{L}_{S_{mean}} = \mathbb{E}_{x_{1:T} \sim p} \left[ \sum_t  \left\| \frac{1}{N} \sum_{n=1}^N \mathbf{S_{t_n}} - \frac{1}{N} \sum_{n=1}^N \mathbf{\tilde{S}_{t_n}} \right\|_2 \right]
\end{equation}

\begin{align}
\label{equ:thirteen}
\mathcal{L}_{S_{std}} = & \mathbb{E}_{x_{1:T} \sim p} \Bigg[ \sum_t \Bigg\| \sqrt{\frac{1}{N} \sum_{n=1}^N (\mathbf{S_{t_n}} - \overline{\mathbf{S_t}})^2} \nonumber \\
& - \sqrt{\frac{1}{N} \sum_{n=1}^N (\mathbf{\tilde{S}_{t_n}} - \overline{\mathbf{\tilde{S}_t}})^2} \Bigg\|_2 \Bigg]
\end{align}

Other components of \(\mathcal{L}_{TS}\), such as skewness, weighted average, and median, are calculated similarly to \eqref{equ:ten}, \eqref{equ:twelve}, and \eqref{equ:thirteen}. The only difference is that instead of using the formula for slope, the formulas for skewness (skew), weighted average (wAvg), and median are applied.

\begin{equation}
\text{skew} = \frac{1}{T} \sum_{t=1}^T \left( \frac{x_t - \bar{x}}{\sigma_x} \right)^3
\end{equation}

\begin{equation}
\text{wAvg} = \frac{\sum_{t=1}^T w_t x_t}{\sum_{t=1}^T w_t}
\end{equation}

Where $\sigma_x$ represents the std of $x$, and $w_t$ denotes the weight assigned to the value $x_t$ at timestamp $t$.

\subsection{GRU-LSTM Network Architecture}

Leveraging the strengths of different neural network architectures by combining them has long been a powerful and effective approach. In auditory attention detection (AAD), combining GRU and CNN architectures has been particularly effective. CNNs, while good at extracting spatial features from EEG data, struggle to capture long-term dependencies. To address this, the AAD-GCQL model \cite{eskandarinasab2024grucnn} integrates GRU with CNN to capture both spatial and temporal dynamics in EEG signals, enhancing the detection of auditory attention.

The GRU used in this combination belongs to a broader family of recurrent neural networks (RNNs), which are tailored for sequence modeling tasks. Among these, LSTM and GRU are the two most prominent architectures, frequently applied in domains such as natural language processing \cite{nlp} and time series forecasting. LSTMs are equipped with memory cells and three distinct gates (input, output, and forget), which help manage the flow of information and address the vanishing gradient problem seen in traditional RNNs \cite{hochreiter1997long}. This architecture makes LSTMs particularly well-suited for longer sequence data, where maintaining information over extended intervals is critical. On the other hand, GRUs simplify the structure by merging the input and forget gates into a single update gate, complemented by a reset gate that determines the extent of past information retention \cite{cho2014learning}. GRUs tend to be more efficient and quicker to train, making them ideal for tasks with shorter sequences or when computational resources are limited. The decision between using LSTM and GRU often hinges on the specific sequence length and complexity of the task, with LSTMs generally preferred for longer sequences and GRUs for shorter ones \cite{lstmgru}.

\begin{figure}
  \centering
  \includegraphics[width=0.5\textwidth]{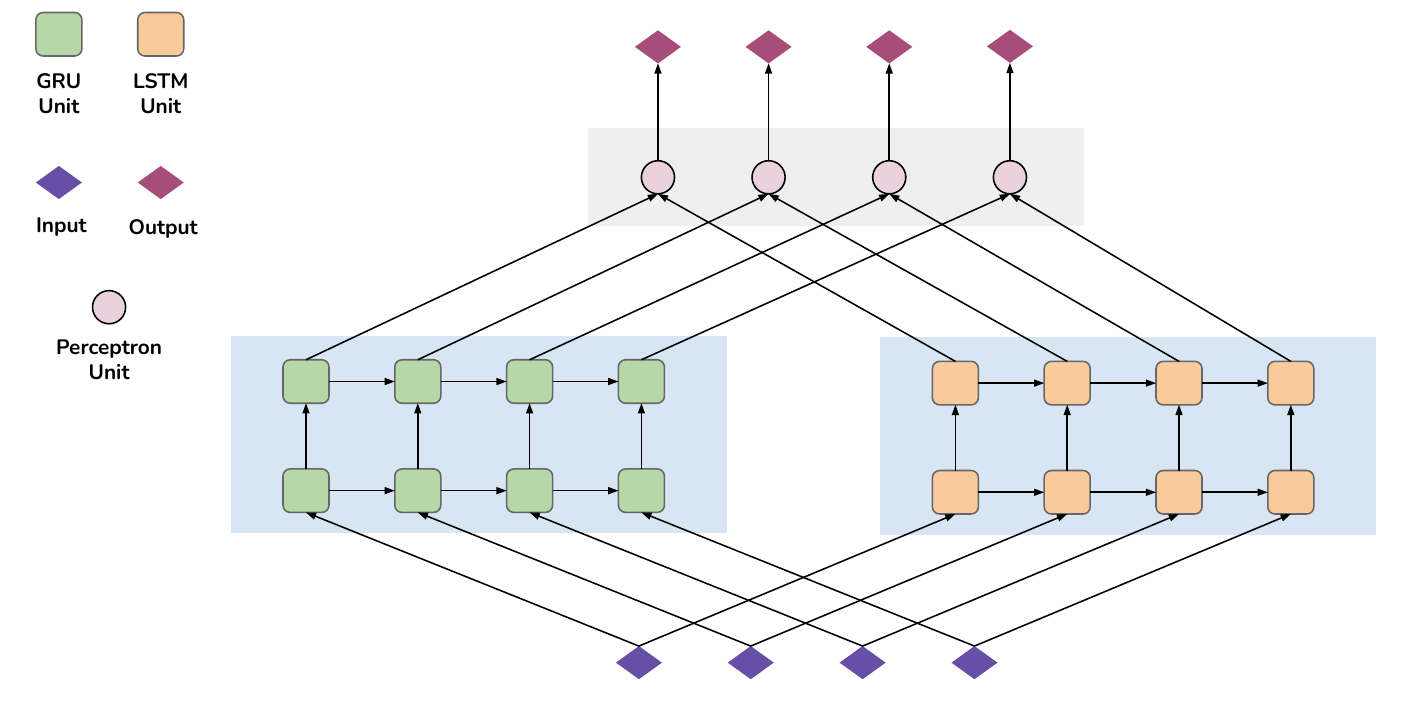}
  \caption{GRU-LSTM Network Architecture: The figure illustrates the architecture of a GRU-LSTM model for univariate time series data, featuring multiple layers of LSTM and GRU cells (in this case, two layers) trained separately. These layers are then combined through perceptron or fully connected neural network layers. For multivariate time series data, multiple instances of these components are trained in parallel.}
  \label{fig:arch}
\end{figure}

A time series generation framework should be capable of handling both short and long sequences and, more importantly, be accurate on both. The exclusive use of either LSTM or GRU as the network architecture can lead to weaknesses in handling either long or short sequences. As shown in Fig. \ref{fig:arch}, by implementing both network architectures and merging the results via a multilayer perceptron, the network becomes more generalized, making it more powerful in learning both long and short sequences. We employ multiple layers of GRU and LSTM separately to produce output, and then merge them using a multilayer perceptron network to obtain the final output. We utilize the same architecture and number of layers for all five networks within the ChronoGAN framework.

\subsection{Early Generation}

Another prevalent issue with GANs is stability. To enhance the stability of the network, we employ an early generation algorithm since the optimal results may be achieved after a random, rather than a specific, number of iterations. Accordingly, as per Algorithm \ref{alg:earlystopping}, after half the number of epochs, we generate synthetic data and calculate the discriminative score and predictive score between real and synthetic data at intervals of every 500 epochs. Additionally, we compute the MSE of the mean and MSE of the std of real and synthetic data to verify whether the synthetic data matches the distribution of the real data. By integrating the results of the discriminative score, predictive score, and MSE of the mean and std, we determine whether to save the current model and generated data. Upon the completion of training, we ensure that the framework has produced the optimal results, consistently delivering reliable and precise outcomes after each training session. It is crucial to determine the appropriate weights for these metrics in order to integrate them and compare them with the previously saved model. The proportion of the discriminative score, predictive score, and MSE of the mean and std can vary depending on the characteristics of the dataset. Therefore, it is inappropriate to establish fixed hyperparameters to combine these three metrics. To address this issue, we initially calculate the hyperparameters \(p1\) and \(p2\) during the first assessment of these metrics. Once established, these hyperparameters are consistently applied in all subsequent epochs.

\begin{algorithm}
\caption{Early Generation Algorithm}
\label{alg:earlystopping}
\small
\begin{algorithmic}
\STATE Initialize $real$ and $synthetic$ samples
\STATE Set $N$ as the total number of epochs
\STATE Initialize $totalError$, $p1$, and $p2$ to None
\STATE Set $checkEpoch \gets 500$ and $startEpoch \gets \lfloor \frac{N}{2} \rfloor$

\FOR {$epoch = 1$ \TO $N$}
    \IF {$epoch \geq startEpoch$ \AND $epoch \bmod checkEpoch == 0$}
        \STATE $disScore \gets \text{calcDis}(real, synthetic)$
        \STATE $preScore \gets \text{calcPre}(real, synthetic)$
        \STATE $meanReal \gets \text{calcMean}(real)$
        \STATE $meanSynth \gets \text{calcMean}(synthetic)$
        \STATE $mseMean \gets \text{calcMSE}(meanReal, meanSynth)$
        
        \STATE $varReal \gets \text{calcVar}(real)$
        \STATE $varSynth \gets \text{calcVar}(synthetic)$
        \STATE $mseVar \gets \text{calcMSE}(varReal, varSynth)$
        \STATE $mseSTD \gets \sqrt{mseVar}$
        \vspace{2mm}

        \IF {$p1 == \text{None}$ \AND $p2 == \text{None}$}
            \STATE $p1 \gets \frac{disScore}{preScore}$
            \vspace{2mm}
            \STATE $p2 \gets \frac{disScore}{mseMean + mseSTD}$
            \vspace{2mm}
        \ENDIF
        \STATE $score \gets disScore + p1 * preScore + p2 * (mseMean + mseSTD)$
        
        \IF {$score \leq totalError$ \OR $totalError == \text{None}$}
            \STATE $totalError \gets score$
            \STATE $\text{saveSynthetic}(synthetic)$
        \ENDIF
    \ENDIF
\ENDFOR
\end{algorithmic}
\end{algorithm}

\section{Experimnets}

The codebase for the ChronoGAN framework, along with a detailed tutorial on its usage, implementation, and hyperparameter settings, is publicly available for review and application \footnote{The codebase of ChronoGAN is available here: \href{https://github.com/samresume/ChronoGAN}{https://github.com/samresume/ChronoGAN}}. The framework is designed to be straightforward, allowing users to simply call a Python function and provide the necessary data and hyperparameters.

\subsection{Datasets}
We evaluate ChronoGAN's effectiveness on time-series datasets with varying attributes such as periodicity, discreteness, noise levels, length, and feature correlation over time. We choose the datasets based on different combinations of these characteristics:

\begin{enumerate}

\item \textbf{Stocks:} Stock price sequences are continuous but aperiodic and features are correlated. We use daily historical data from Google stocks spanning 2004 to 2019, which includes features such as volume, high, low, opening, closing, and adjusted closing prices.

\item \textbf{Sines:} We generate multivariate sinusoidal sequences with varying frequencies $\eta$ and phases $\theta$, providing continuous, periodic, and multivariate data with each feature being independent.

\item \textbf{ECG:} The ECG5000 dataset from Physionet, which covers a 20-hour long ECG recording with 140 timestamps, is a univariate time series that is continuous and periodic. The data is classified as a long time series.

\item \textbf{SWAN-SF:} The Space Weather Analytics for Solar Flares (SWAN-SF) \cite{swansf} dataset consists of multivariate time series of photospheric magnetic field parameters for solar flare prediction tasks \cite{sfpredict}. The SWAN-SF dataset is recognized as challenging due to its complex temporal dynamics and the numerous data preprocessing issues it presents. In \cite{eskandari_apjs}, the authors thoroughly addressed these challenges by implementing an innovative preprocessing pipeline \cite{notebooks}. This effort resulted in the creation of an enhanced version of the SWAN-SF dataset \cite{clean_swansf}, which was subsequently utilized in our evaluation in place of the original, unprocessed dataset.
\end{enumerate}

\subsection{Baseline Techniques and Evaluation Metrics}

We conduct a comparison between ChronoGAN, TimeGAN \cite{NEURIPS2019_c9efe5f2}, Teacher Forcing (T-Forcing) \cite{Tforcing}, Professor Forcing (P-Forcing) \cite{Pforcing} and Standard GAN \cite{esteban2017realvalued}, which represent the five best-performing techniques in various fields of time series generation, including GAN-based and Autoregressive approaches. To ensure unbiased results, we maintain identical hyperparameters across all five models. To evaluate the quality of the generated data, we focus on three key criteria:

\begin{enumerate}
  \item \textbf{Visualization}: We utilize t-SNE \cite{JMLR:v9:vandermaaten08a} and PCA \cite{bryant1995principal} analyses on both the original and synthetic datasets. This approach aids in qualitatively assessing how closely the distribution of the generated samples matches that of the original in a two-dimensional space.
  
  \item \textbf{Discriminative Score}: For a quantitative measure of similarity, each sequence from the original dataset is labeled as `real`, while each from the generated set is labeled as `synthetic`. An LSTM classifier is then trained to differentiate these two categories in a standard supervised learning task. The classification error on a reserved test set provides a quantitative measure of this score. We then subtract the result from 0.5, making the optimal result 0 instead of 0.5 for easier comparison.
  
  \item \textbf{Predictive Score}: To evaluate the quality of the generated data in capturing step-wise conditional distributions, we utilize the synthetic dataset to train an LSTM for sequence prediction. This involves forecasting the next-step temporal vectors for each input sequence. The model’s accuracy is subsequently tested on the original dataset, with performance assessed using the MAE.
\end{enumerate}

For each discriminative or predictive score experiment, we replicated the experiments eight times to avoid incidental results. We present the mean and std of each experiment in Tables \ref{tbl:discriminative} and \ref{tbl:predictive}.

\begin{table}
\centering
\caption{Comparative analysis of discriminative score for leading time series generation techniques (Lower scores are better)}
\label{tbl:discriminative}
\scriptsize
\begin{tabular}{|c|c|c|c|c|}
\hline
\textbf{} & \textbf{Stocks} & \textbf{Sines} & \textbf{ECG} & \textbf{SWAN-SF} \\ \hline

\textbf{ChronoGAN} & \textbf{0.204 \raisebox{-2.5pt}{\stackon[-2pt]{$-$}{$+$}} 0.03} & \textbf{0.190 \raisebox{-2.5pt}{\stackon[-2pt]{$-$}{$+$}} 0.08} & \textbf{0.213 \raisebox{-2.5pt}{\stackon[-2pt]{$-$}{$+$}} 0.04} & \textbf{0.304 \raisebox{-2.5pt}{\stackon[-2pt]{$-$}{$+$}} 0.06} \\ \hline

\textbf{TimeGAN} & 0.326 \raisebox{-2.5pt}{\stackon[-2pt]{$-$}{$+$}} 0.03 & 0.283 \raisebox{-2.5pt}{\stackon[-2pt]{$-$}{$+$}} 0.13 & 0.271 \raisebox{-2.5pt}{\stackon[-2pt]{$-$}{$+$}} 0.08 & 0.374 \raisebox{-2.5pt}{\stackon[-2pt]{$-$}{$+$}} 0.10 \\ \hline

\textbf{GAN} & 0.499 \raisebox{-2.5pt}{\stackon[-2pt]{$-$}{$+$}} 0.01 & 0.320 \raisebox{-2.5pt}{\stackon[-2pt]{$-$}{$+$}} 0.22 & 0.486 \raisebox{-2.5pt}{\stackon[-2pt]{$-$}{$+$}} 0.01 & 0.5 \raisebox{-2.5pt}{\stackon[-2pt]{$-$}{$+$}} 0.00 \\ \hline

\textbf{T-Forcing} & 0.476 \raisebox{-2.5pt}{\stackon[-2pt]{$-$}{$+$}} 0.01 & 0.348 \raisebox{-2.5pt}{\stackon[-2pt]{$-$}{$+$}} 0.13 & 0.351 \raisebox{-2.5pt}{\stackon[-2pt]{$-$}{$+$}} 0.10 & 0.5 \raisebox{-2.5pt}{\stackon[-2pt]{$-$}{$+$}} 0.00 \\ \hline

\textbf{P-Forcing} & 0.5 \raisebox{-2.5pt}{\stackon[-2pt]{$-$}{$+$}} 0.00 & 0.5 \raisebox{-2.5pt}{\stackon[-2pt]{$-$}{$+$}} 0.00 & 0.329 \raisebox{-2.5pt}{\stackon[-2pt]{$-$}{$+$}} 0.10 & 0.5 \raisebox{-2.5pt}{\stackon[-2pt]{$-$}{$+$}} 0.00 \\ \hline

\end{tabular}
\end{table}

\begin{table}
\centering
\caption{Comparative analysis of predictive score for leading time series generation techniques (Lower scores are better)}
\label{tbl:predictive}
\scriptsize
\begin{tabular}{|c|c|c|c|c|}
\hline

\textbf{} & \textbf{Stocks} & \textbf{Sines} & \textbf{ECG} & \textbf{SWAN-SF} \\ \hline

\textbf{ChronoGAN} & \textbf{0.045 \raisebox{-2.5pt}{\stackon[-2pt]{$-$}{$+$}} 0.00} & \textbf{0.225 \raisebox{-2.5pt}{\stackon[-2pt]{$-$}{$+$}} 0.01} & \textbf{0.129 \raisebox{-2.5pt}{\stackon[-2pt]{$-$}{$+$}} 0.00} & \textbf{0.055 \raisebox{-2.5pt}{\stackon[-2pt]{$-$}{$+$}} 0.00} \\ \hline

\textbf{TimeGAN} & 0.046 \raisebox{-2.5pt}{\stackon[-2pt]{$-$}{$+$}} 0.00 & 0.245 \raisebox{-2.5pt}{\stackon[-2pt]{$-$}{$+$}} 0.01 & 0.129 \raisebox{-2.5pt}{\stackon[-2pt]{$-$}{$+$}} 0.01 & 0.082 \raisebox{-2.5pt}{\stackon[-2pt]{$-$}{$+$}} 0.00 \\ \hline

\textbf{GAN} & 0.186 \raisebox{-2.5pt}{\stackon[-2pt]{$-$}{$+$}} 0.01 & 0.233 \raisebox{-2.5pt}{\stackon[-2pt]{$-$}{$+$}} 0.01 & 0.191 \raisebox{-2.5pt}{\stackon[-2pt]{$-$}{$+$}} 0.00 & 0.219 \raisebox{-2.5pt}{\stackon[-2pt]{$-$}{$+$}} 0.01 \\ \hline

\textbf{T-Forcing} & 0.050 \raisebox{-2.5pt}{\stackon[-2pt]{$-$}{$+$}} 0.01 & 0.275 \raisebox{-2.5pt}{\stackon[-2pt]{$-$}{$+$}} 0.01 & 0.130 \raisebox{-2.5pt}{\stackon[-2pt]{$-$}{$+$}} 0.01 & 0.066 \raisebox{-2.5pt}{\stackon[-2pt]{$-$}{$+$}} 0.01 \\ \hline

\textbf{P-Forcing} & 0.147 \raisebox{-2.5pt}{\stackon[-2pt]{$-$}{$+$}} 0.02 & 0.224 \raisebox{-2.5pt}{\stackon[-2pt]{$-$}{$+$}} 0.01 & 0.194 \raisebox{-2.5pt}{\stackon[-2pt]{$-$}{$+$}} 0.01 & 0.241 \raisebox{-2.5pt}{\stackon[-2pt]{$-$}{$+$}} 0.01 \\ \hline

\end{tabular}
\end{table}

\begin{figure}
  \centering
  \includegraphics[width=0.47\textwidth]{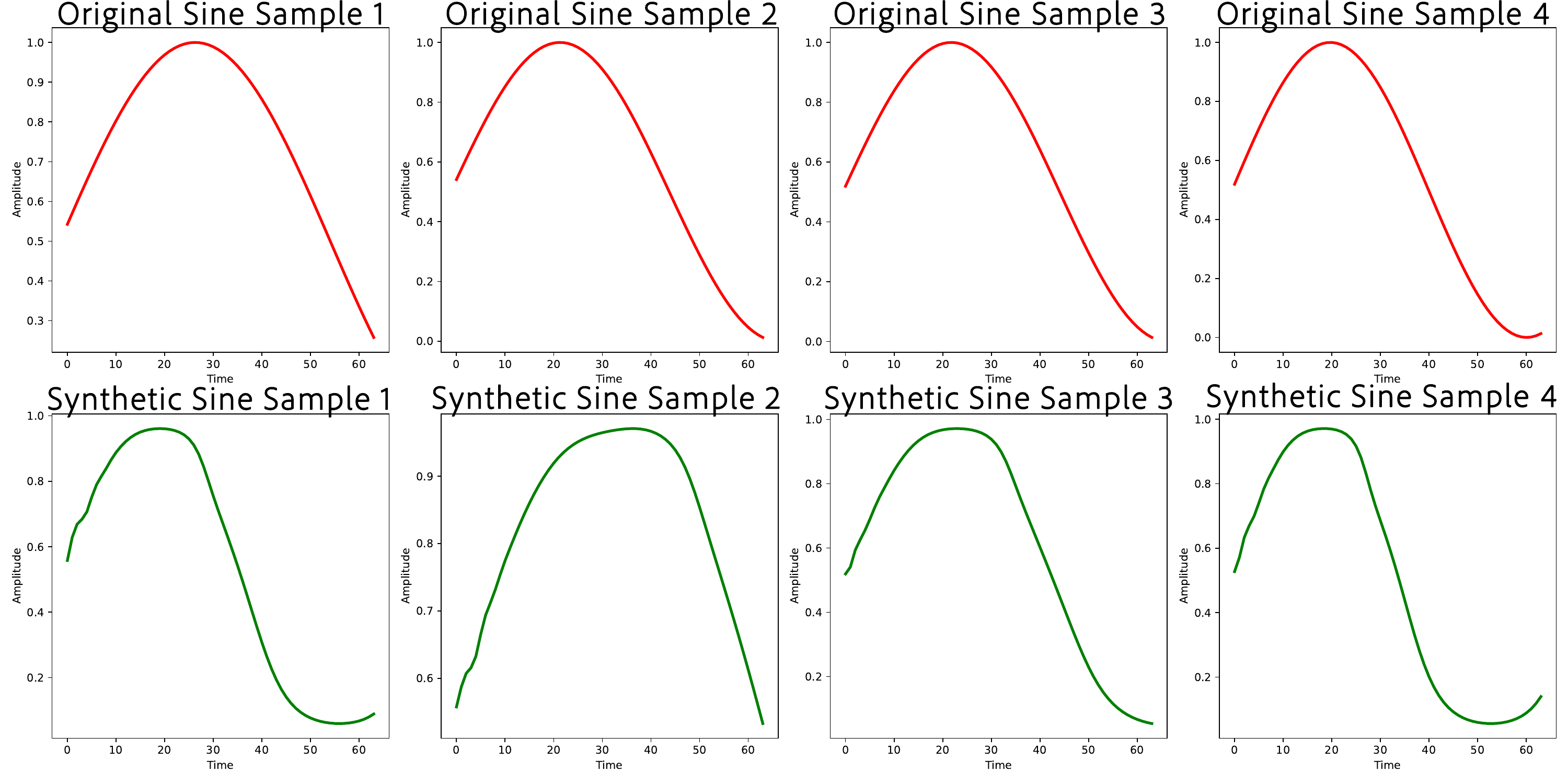}
  \caption{This figure illustrates the original Sines dataset samples (top) and their corresponding synthetic counterparts generated by the ChronoGAN algorithm (bottom). Each subplot shows one of four randomly selected samples.}
  \label{fig:sinesshow}
\end{figure}

\begin{figure}
  \centering
  \includegraphics[width=0.47\textwidth]{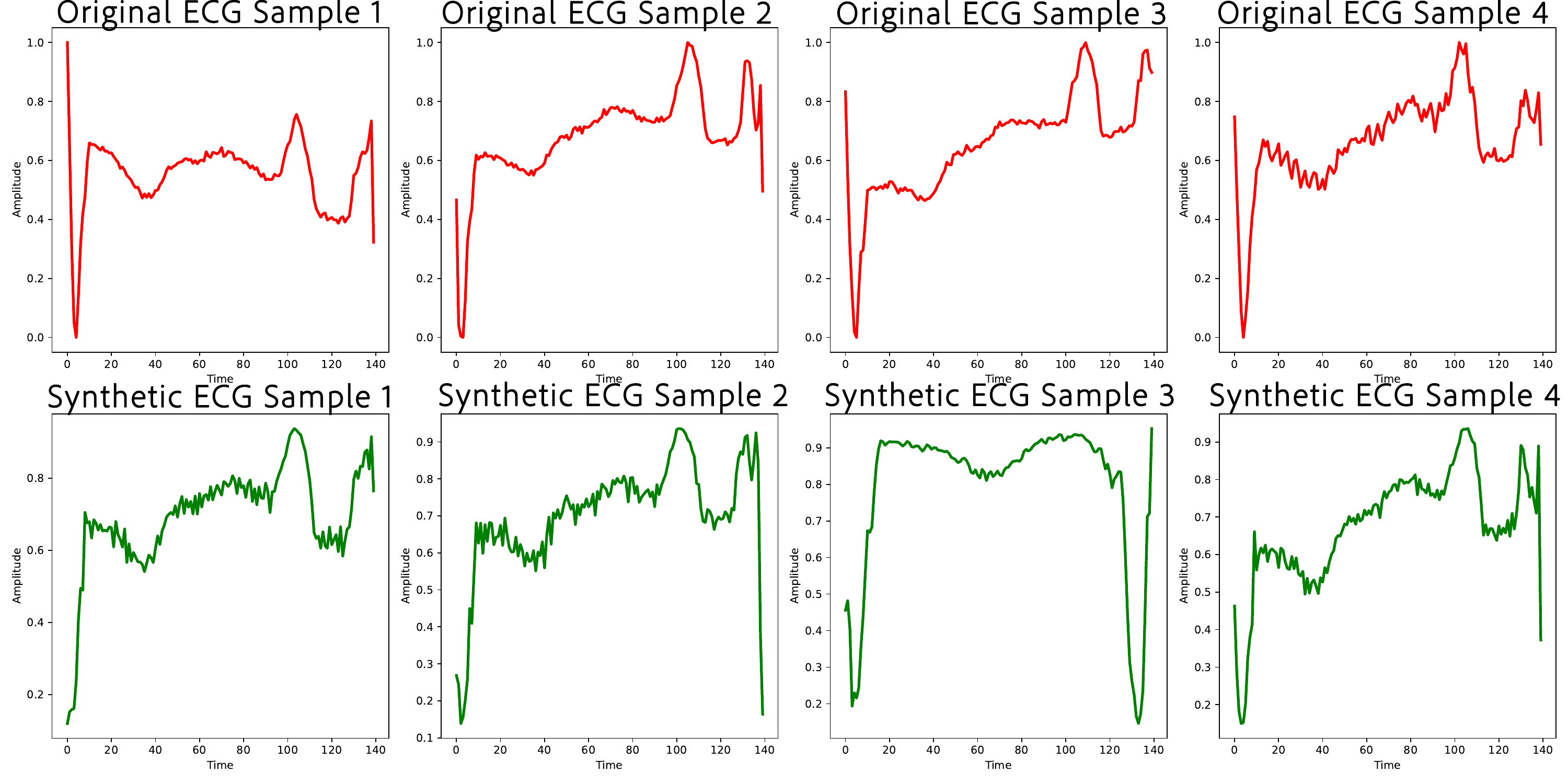}
  \caption{Displayed here are original ECG dataset samples (top) and the synthetic data generated by ChronoGAN (bottom).}
  \label{fig:ecg}
\end{figure}

\subsection{Results and Discussion}

Based on the results presented in Tables \ref{tbl:discriminative} and \ref{tbl:predictive}, the ChronoGAN framework consistently outperforms state-of-the-art models, including TimeGAN, Teacher Forcing, Professor Forcing, and Standard GAN. In terms of the discriminative score, ChronoGAN achieves an average reduction of approximately 27.60\% across the four datasets compared to TimeGAN. This substantial improvement indicates that ChronoGAN generates more realistic temporal data than other techniques. Furthermore, this improvement in the discriminative score can be attributed to the early generation algorithm, which enhances stability and ensures the best data is preserved during training. The improvement is also evident across all four datasets, each with different lengths, demonstrating the effectiveness of the GRU-LSTM layers within our framework. Additionally, according to discriminative score evaluations, ChronoGAN and TimeGAN emerge as superior compared to Teacher Forcing and Standard GAN. This underscores the importance of developing GAN-based techniques specifically tailored for time series data.

In terms of predictive score, ChronoGAN reduces the error by approximately 10.82\% across the four datasets compared to TimeGAN. This underscores the effectiveness of our novel time series-based (\(\mathcal{L}_{TS}\)) and supervised (\(\mathcal{L}_S\)) loss functions, which significantly improve the generator’s ability to capture the temporal dynamics of the data more accurately. As demonstrated in Figs. \ref{fig:sinesshow} and \ref{fig:ecg}, we present several examples of synthetic samples generated by ChronoGAN for both the Sines and ECG datasets. These examples highlight ChronoGAN’s ability to effectively learn the temporal distributions of the real data and generate high-quality synthetic data that accurately reflect those patterns.

\begin{figure}
\centering
\subfloat{\includegraphics[width=0.122\textwidth]{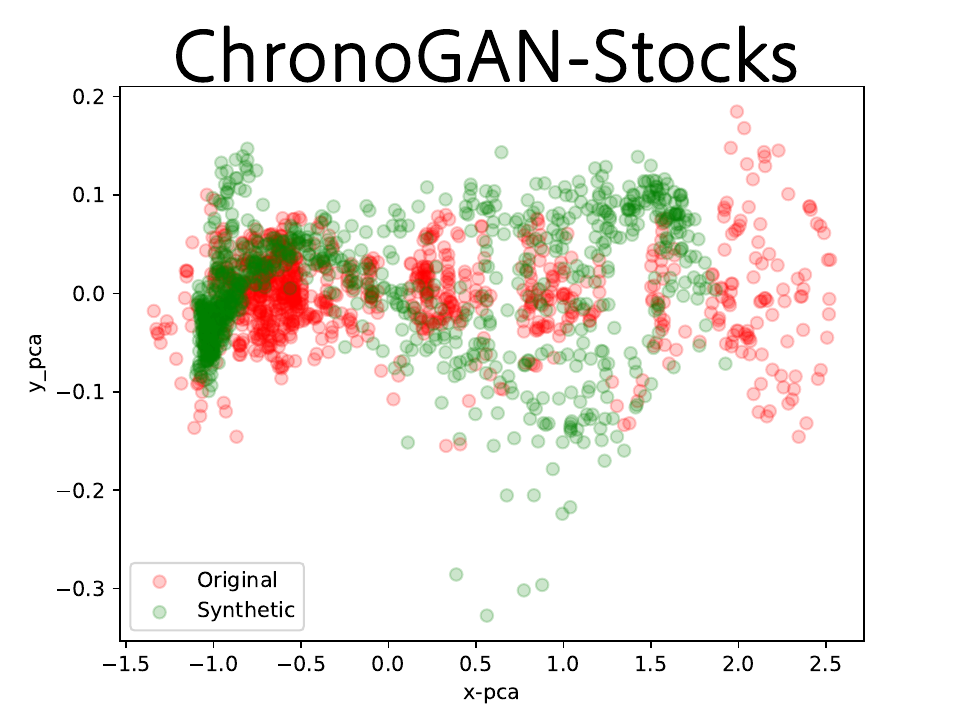}}\hfill
\subfloat{\includegraphics[width=0.122\textwidth]{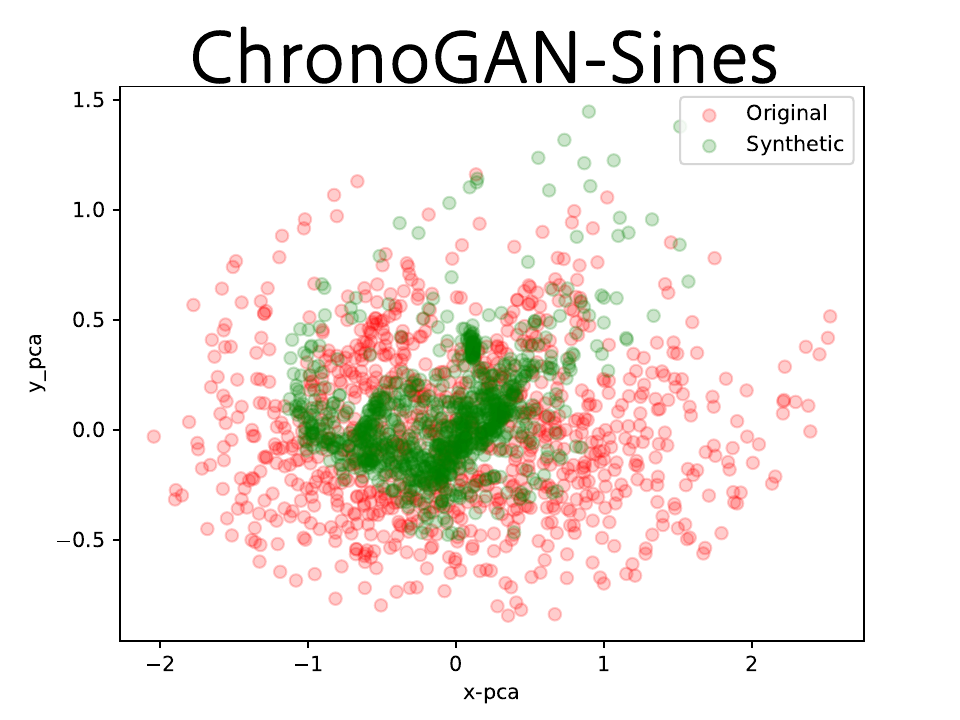}}\hfill
\subfloat{\includegraphics[width=0.122\textwidth]{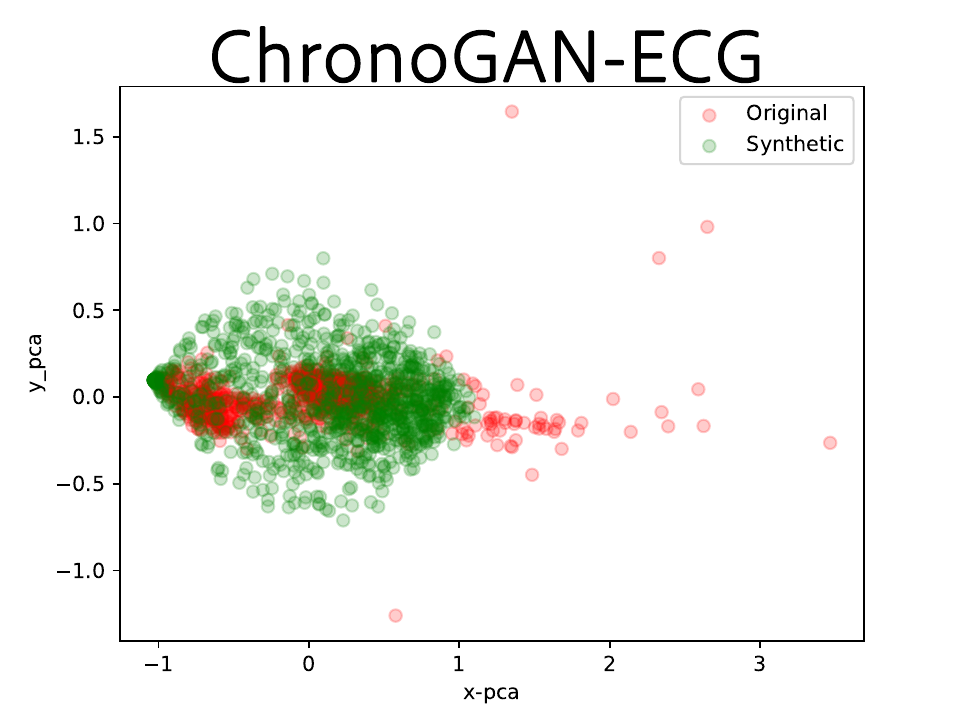}}\hfill
\subfloat{\includegraphics[width=0.122\textwidth]{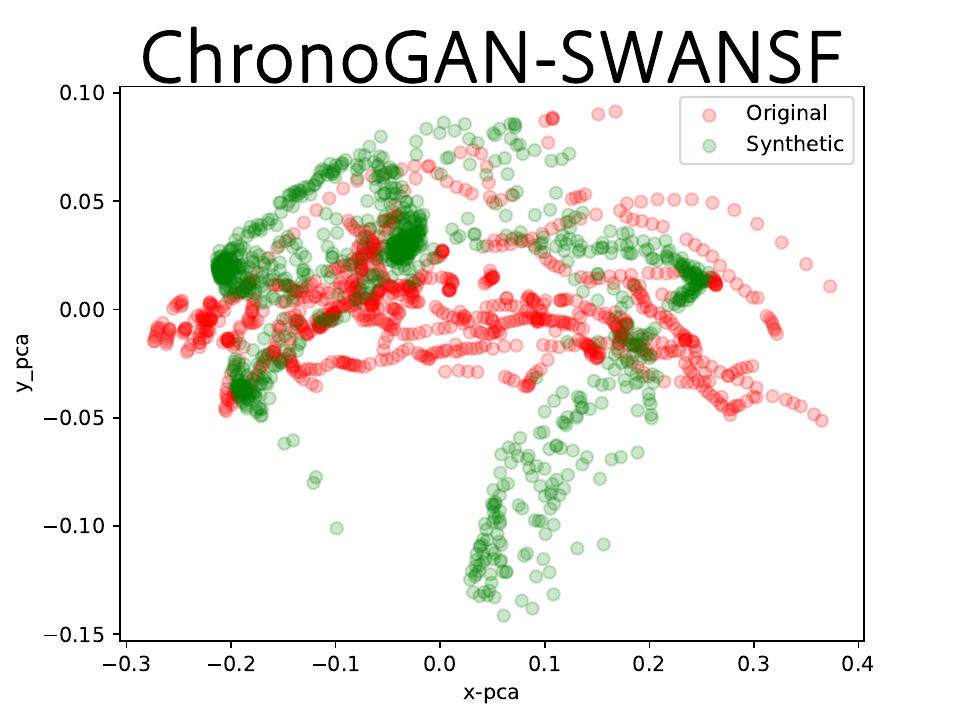}}\hfill

\subfloat{\includegraphics[width=0.122\textwidth]{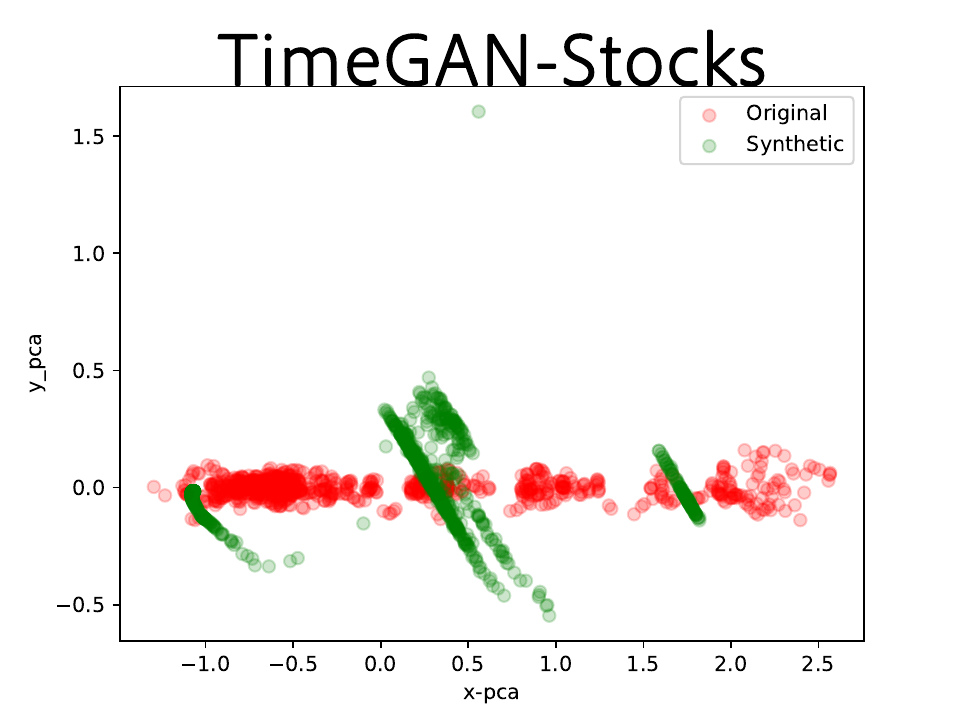}}\hfill
\subfloat{\includegraphics[width=0.122\textwidth]{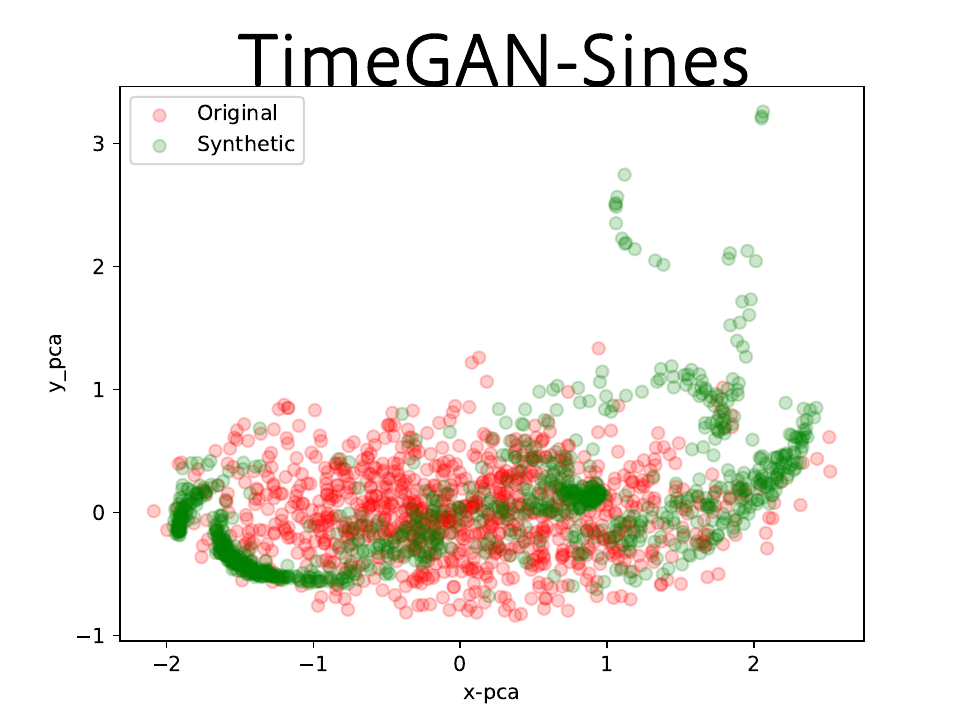}}\hfill
\subfloat{\includegraphics[width=0.122\textwidth]{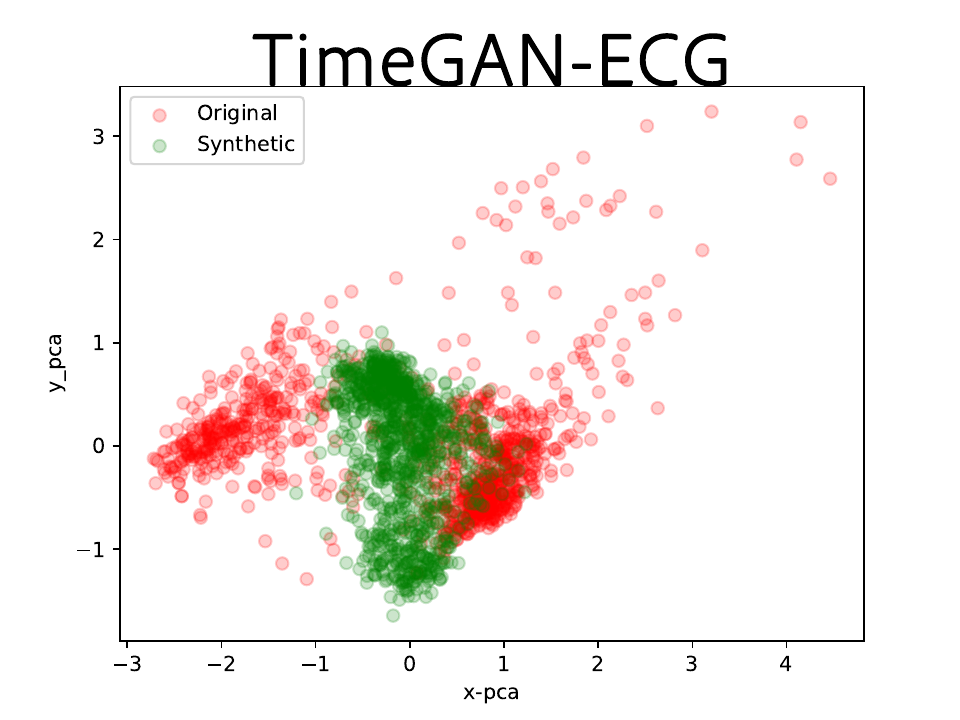}}\hfill
\subfloat{\includegraphics[width=0.122\textwidth]{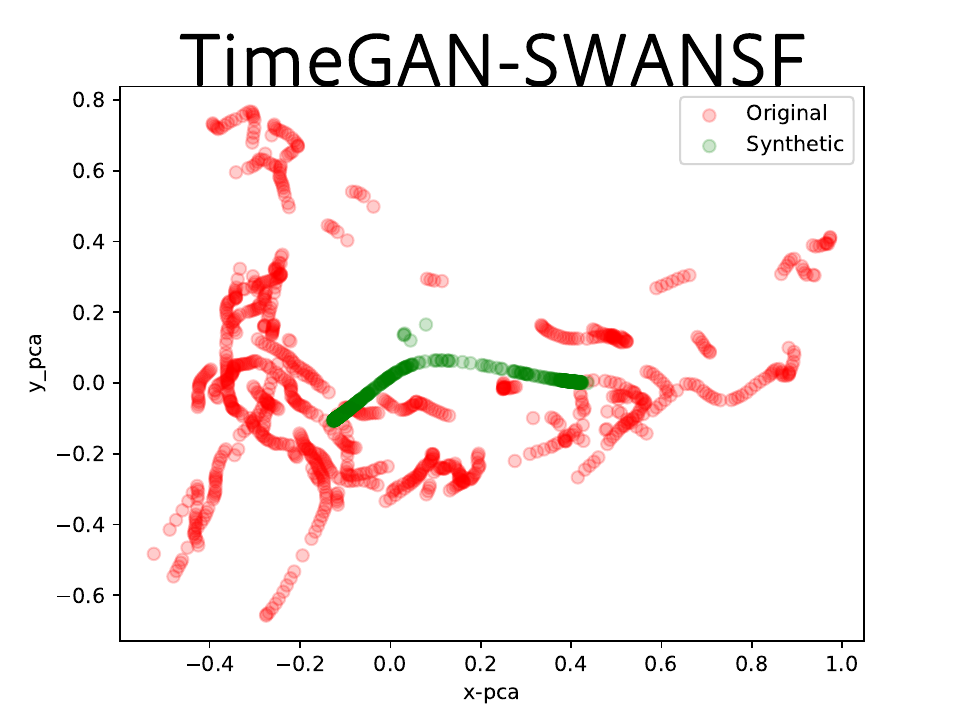}}\hfill

\subfloat{\includegraphics[width=0.122\textwidth]{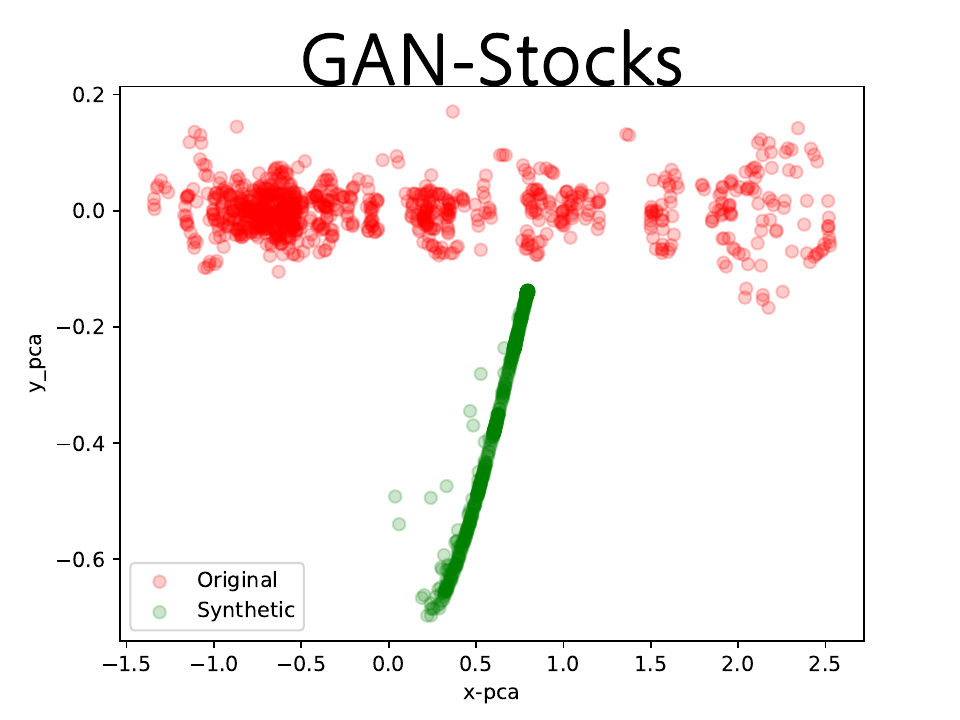}}\hfill
\subfloat{\includegraphics[width=0.122\textwidth]{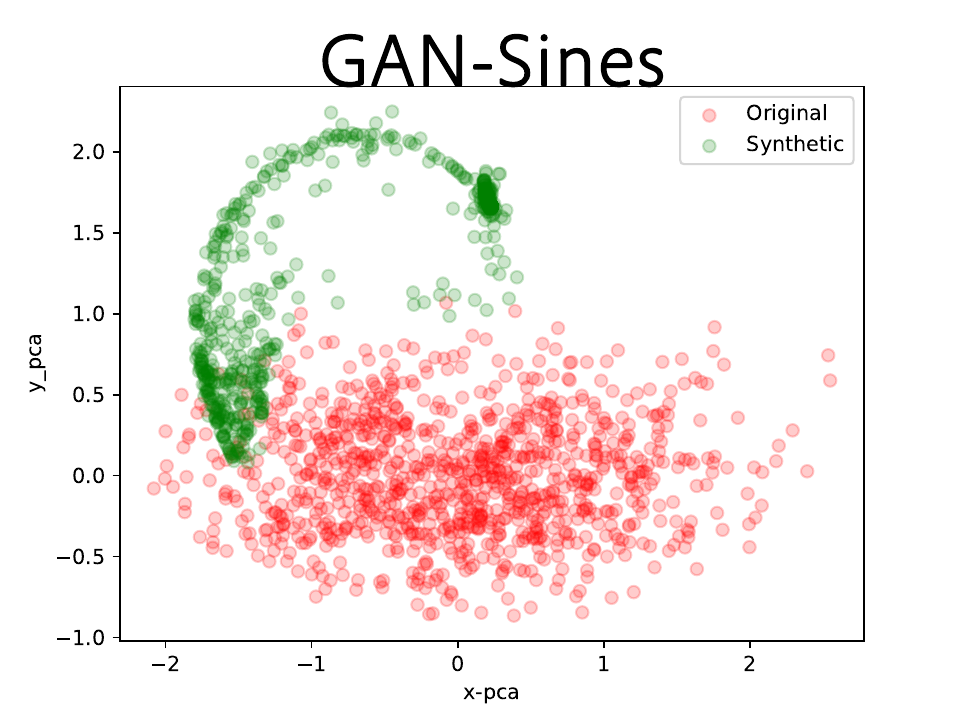}}\hfill
\subfloat{\includegraphics[width=0.122\textwidth]{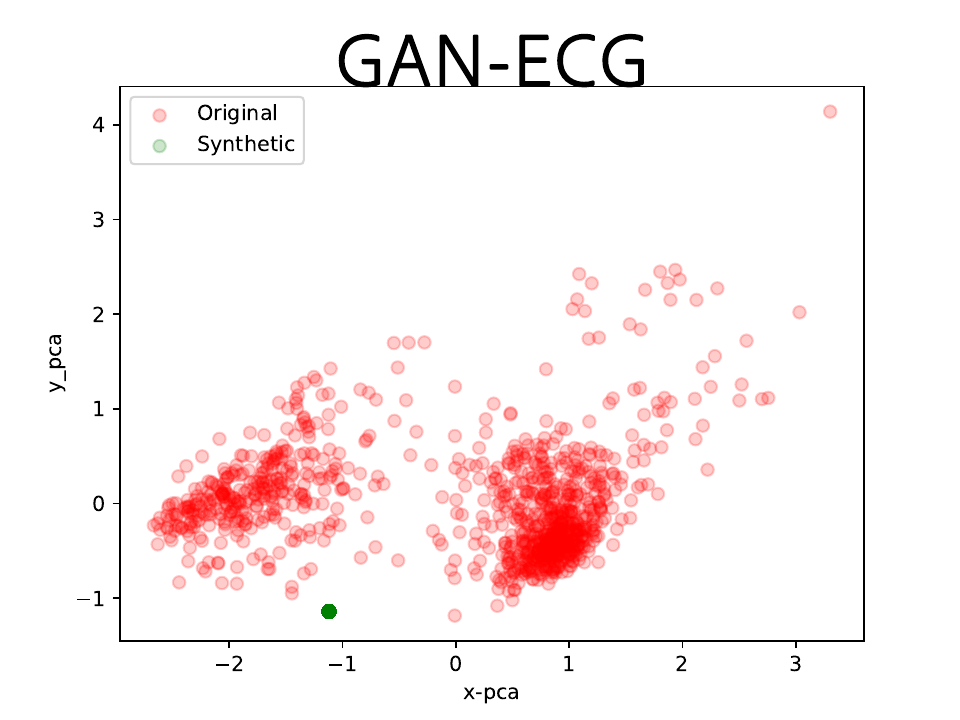}}\hfill
\subfloat{\includegraphics[width=0.122\textwidth]{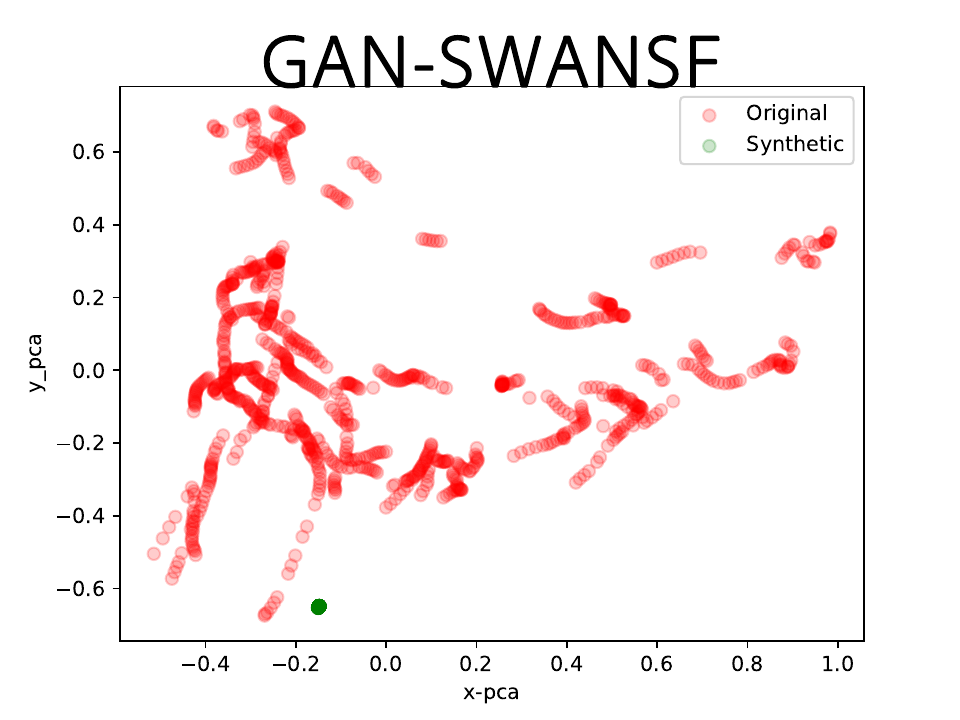}}\hfill

\subfloat{\includegraphics[width=0.122\textwidth]{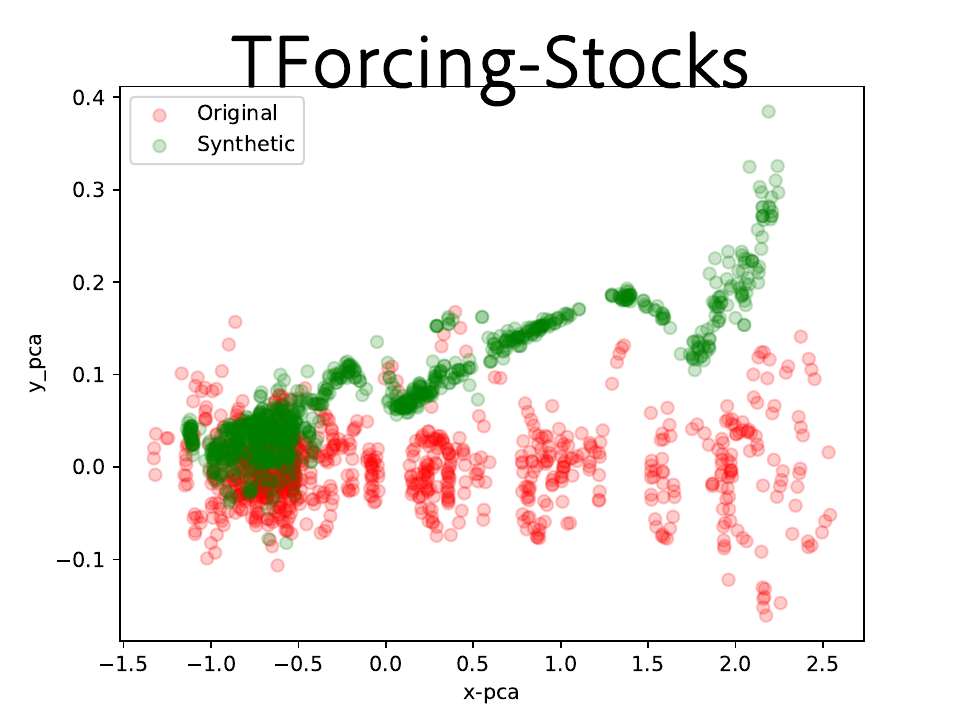}}\hfill
\subfloat{\includegraphics[width=0.122\textwidth]{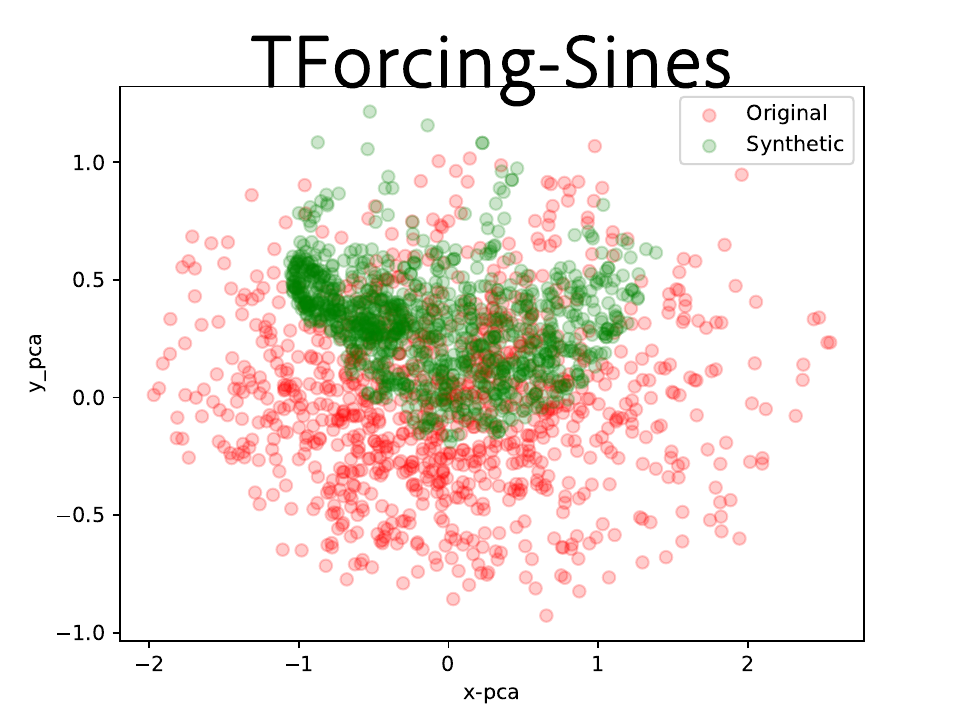}}\hfill
\subfloat{\includegraphics[width=0.122\textwidth]{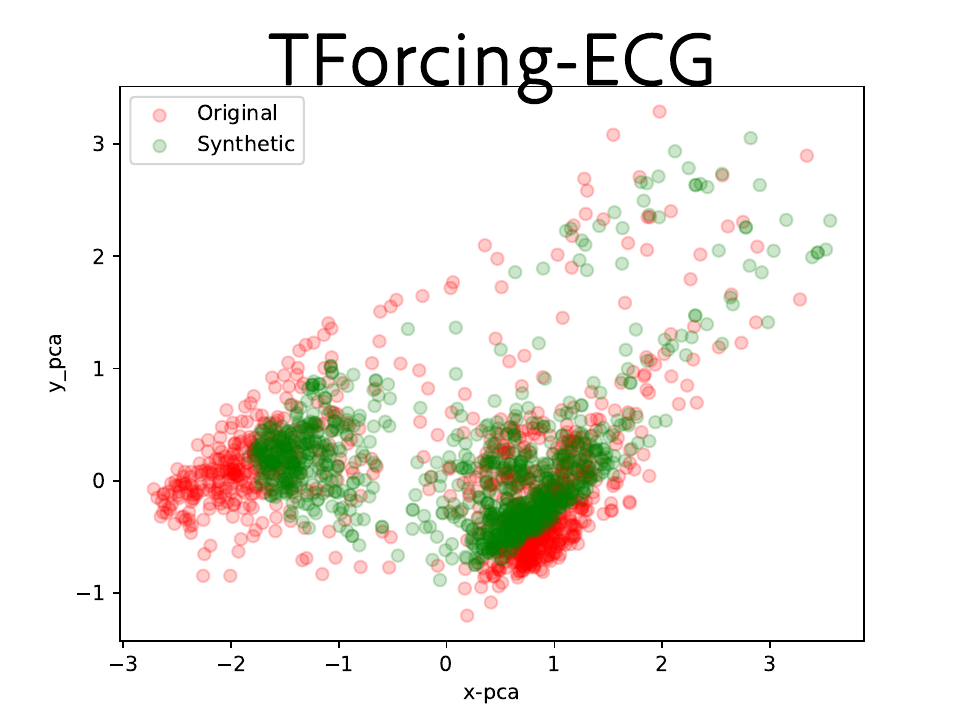}}\hfill
\subfloat{\includegraphics[width=0.122\textwidth]{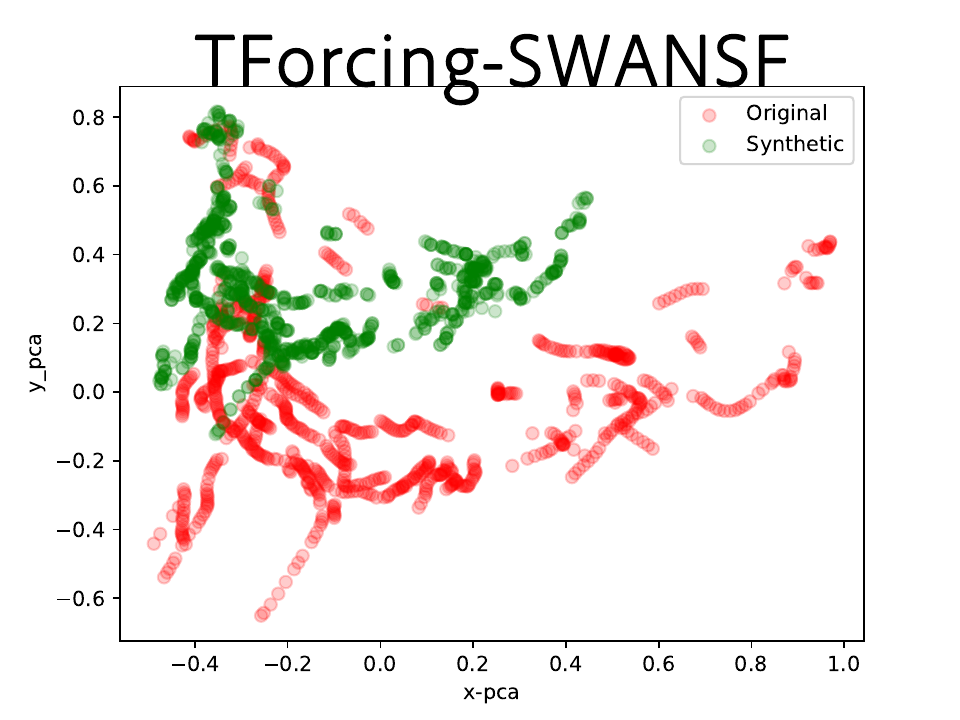}}\hfill

\subfloat{\includegraphics[width=0.122\textwidth]{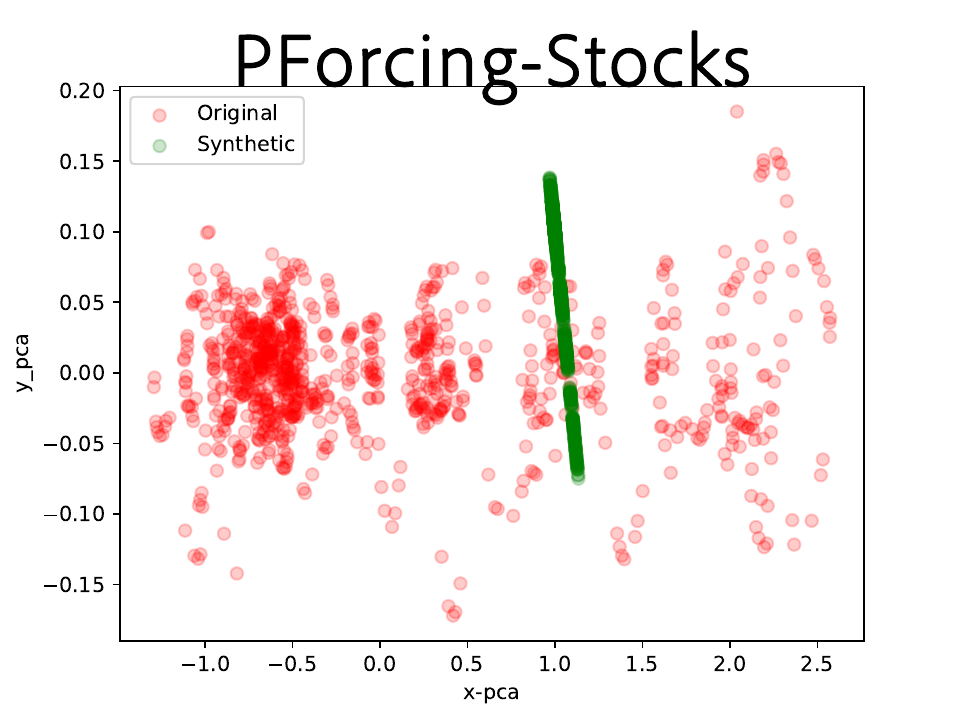}}\hfill
\subfloat{\includegraphics[width=0.122\textwidth]{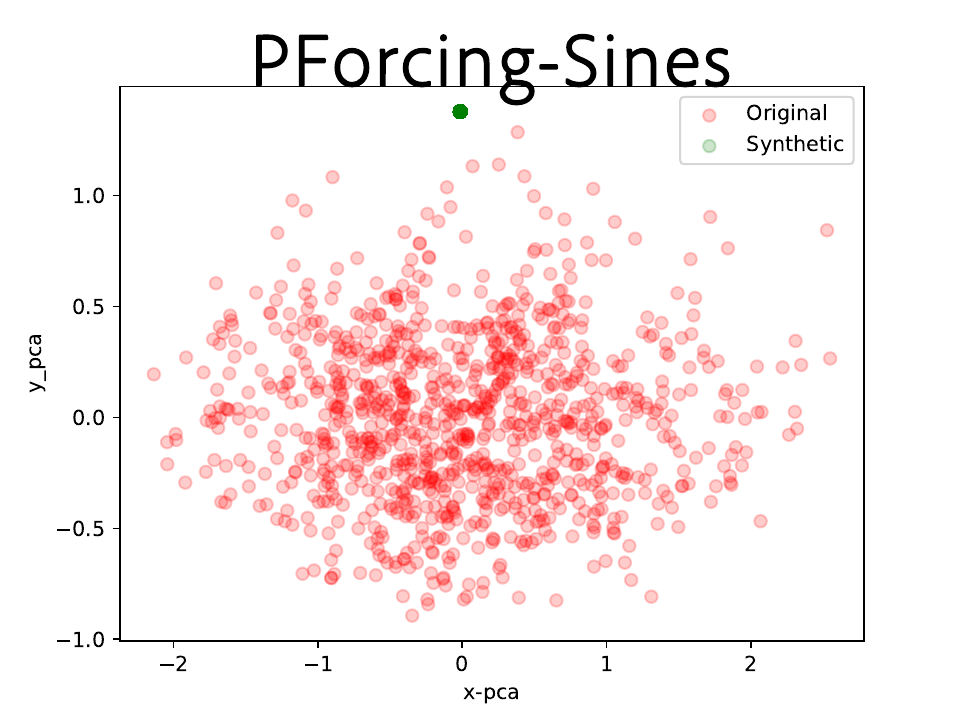}}\hfill
\subfloat{\includegraphics[width=0.122\textwidth]{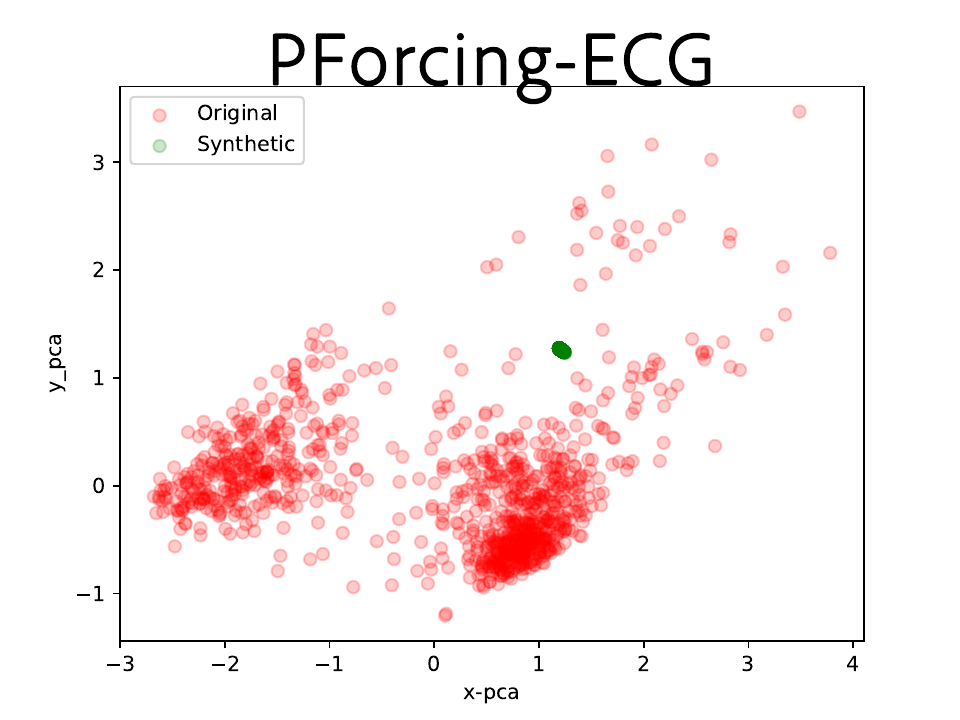}}\hfill
\subfloat{\includegraphics[width=0.122\textwidth]{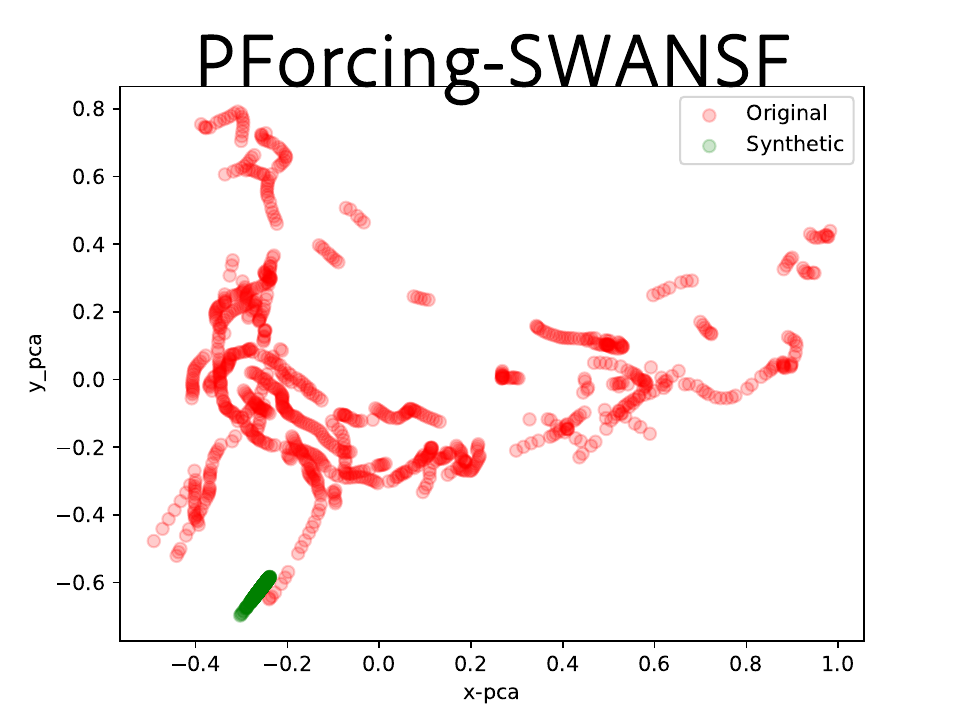}}\hfill

\caption{PCA visualizations illustrate the distributional alignment between original and synthetic data samples generated by ChronoGAN and other baselines across our four datasets.}
\label{fig:stocks}
\end{figure}

\begin{figure}
\centering
\subfloat{\includegraphics[width=0.122\textwidth]{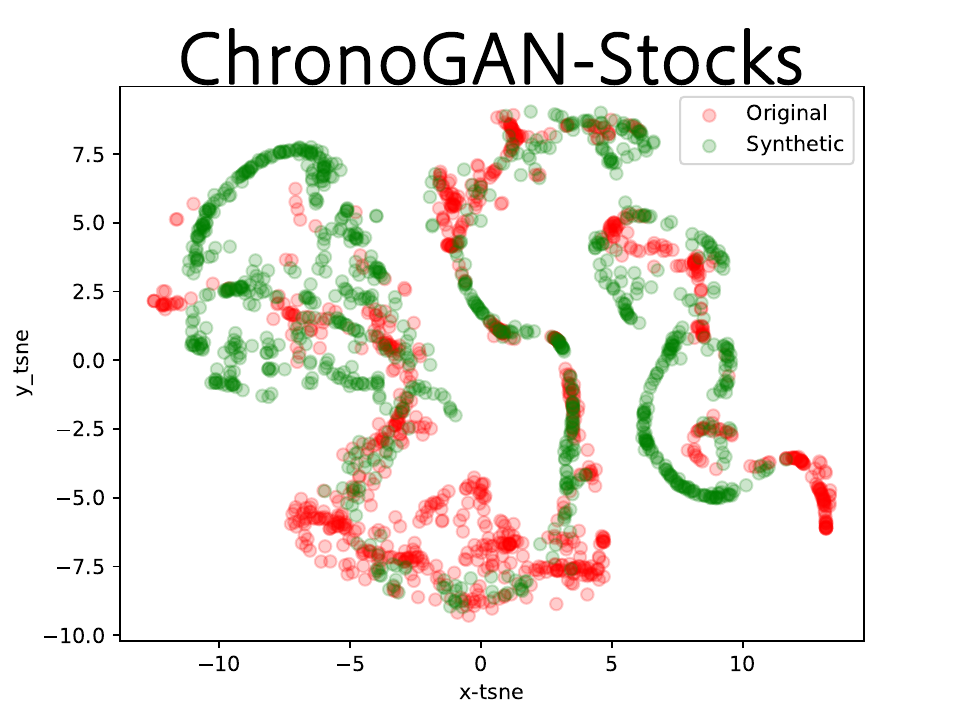}}\hfill
\subfloat{\includegraphics[width=0.122\textwidth]{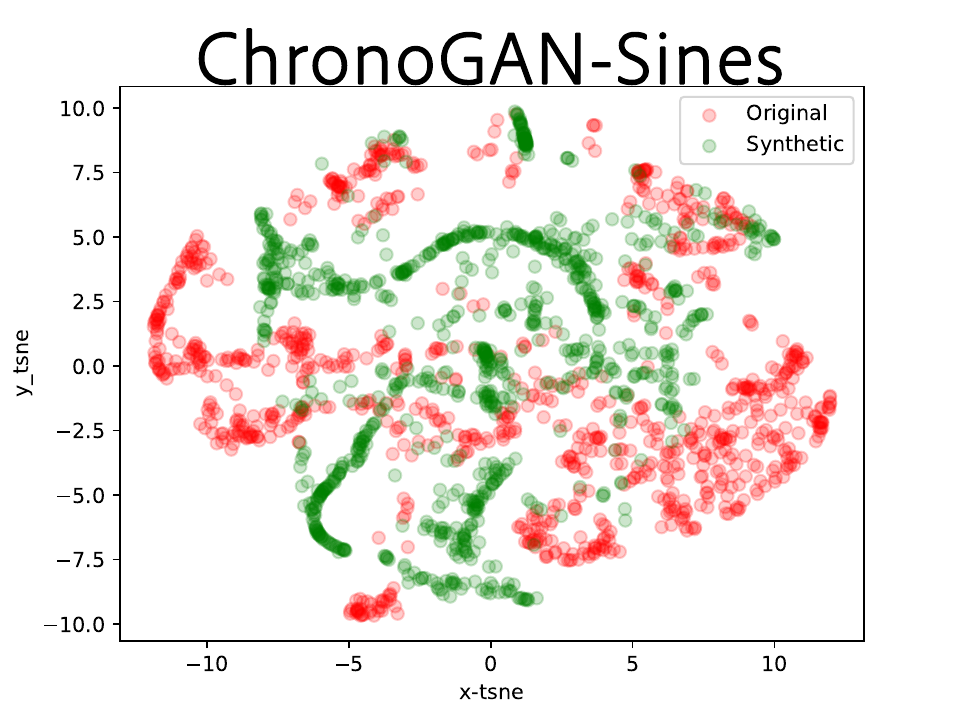}}\hfill
\subfloat{\includegraphics[width=0.122\textwidth]{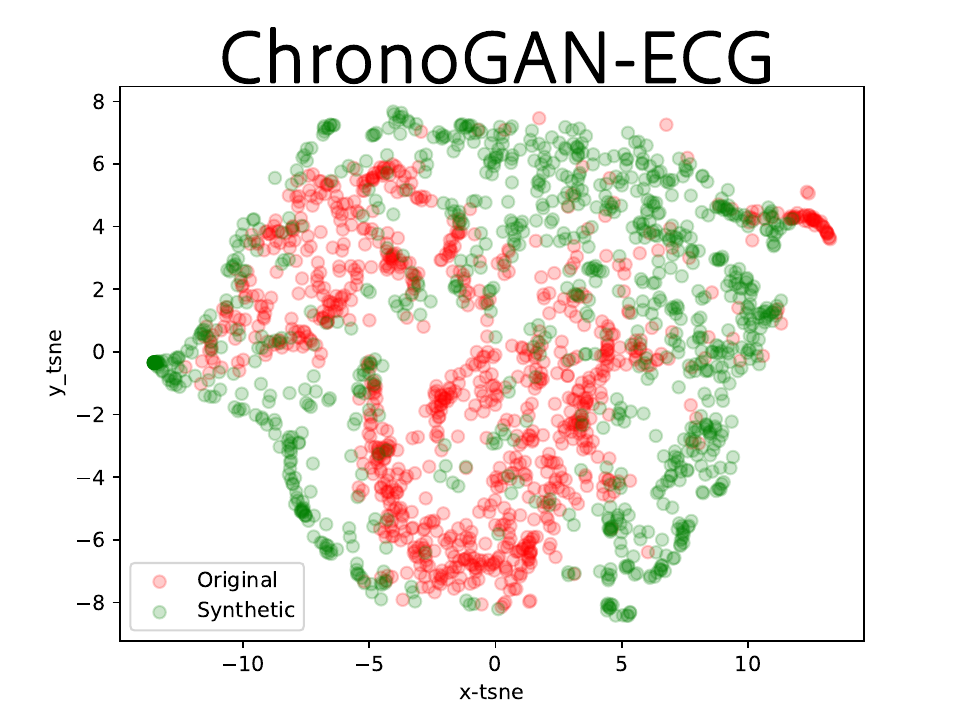}}\hfill
\subfloat{\includegraphics[width=0.122\textwidth]{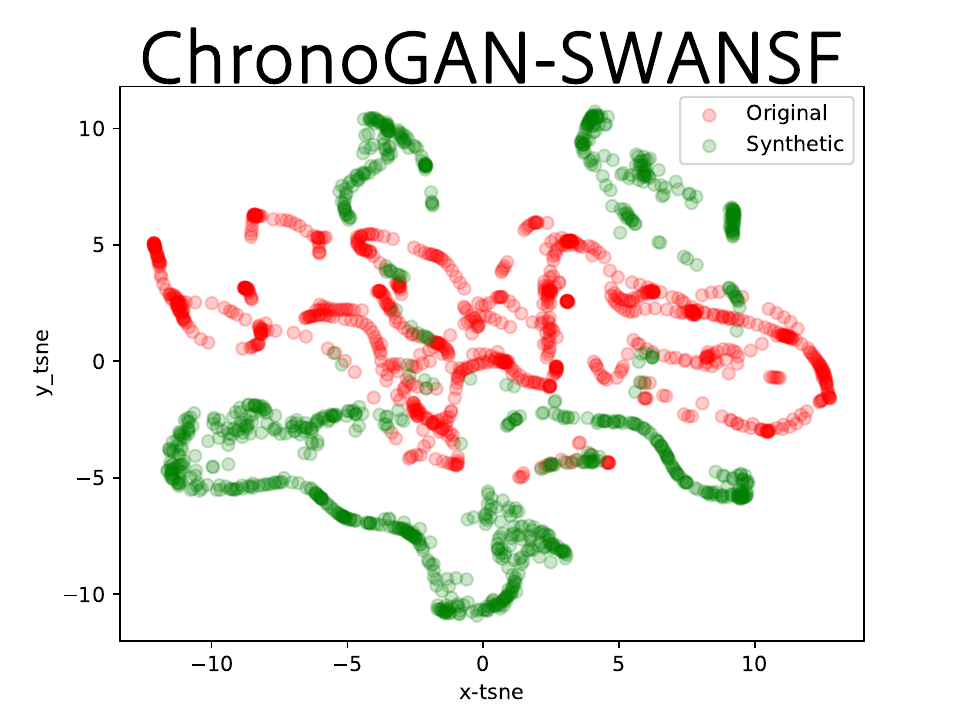}}\hfill

\subfloat{\includegraphics[width=0.122\textwidth]{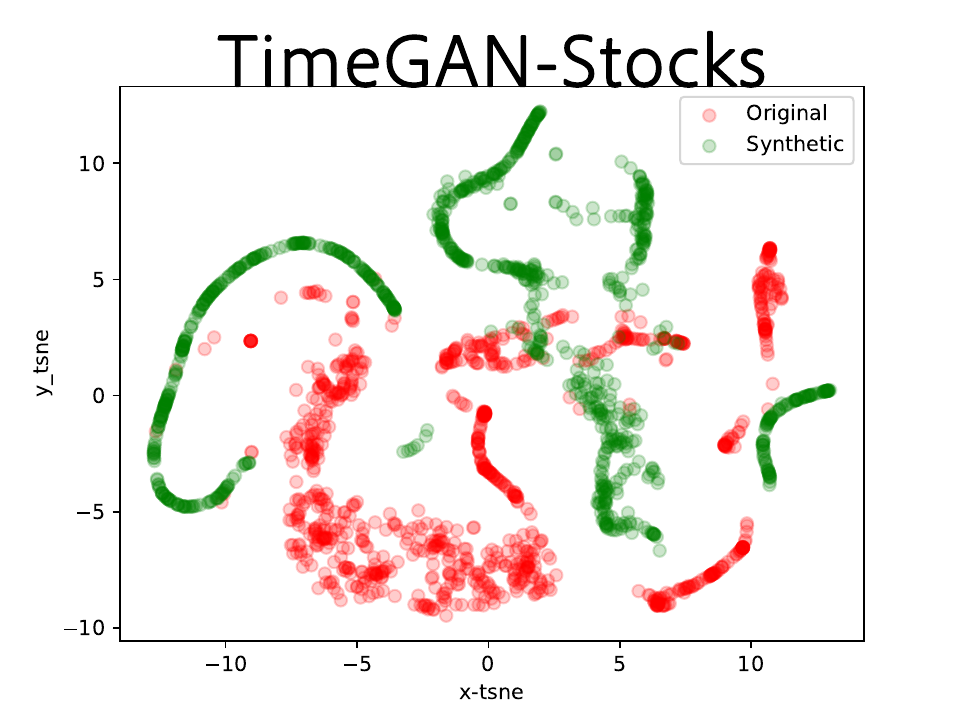}}\hfill
\subfloat{\includegraphics[width=0.122\textwidth]{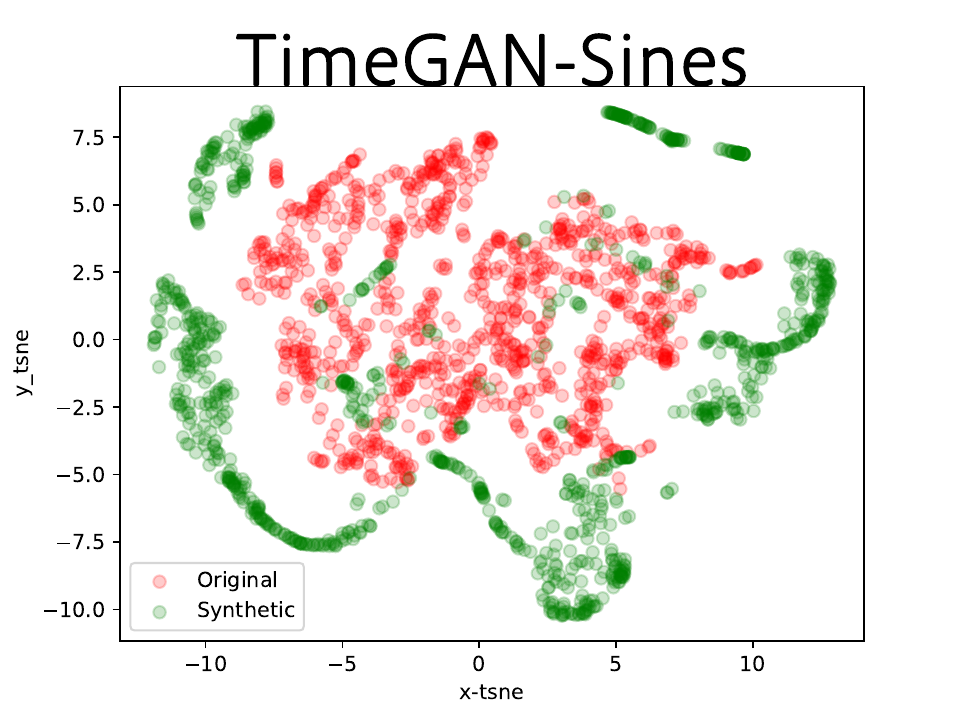}}\hfill
\subfloat{\includegraphics[width=0.122\textwidth]{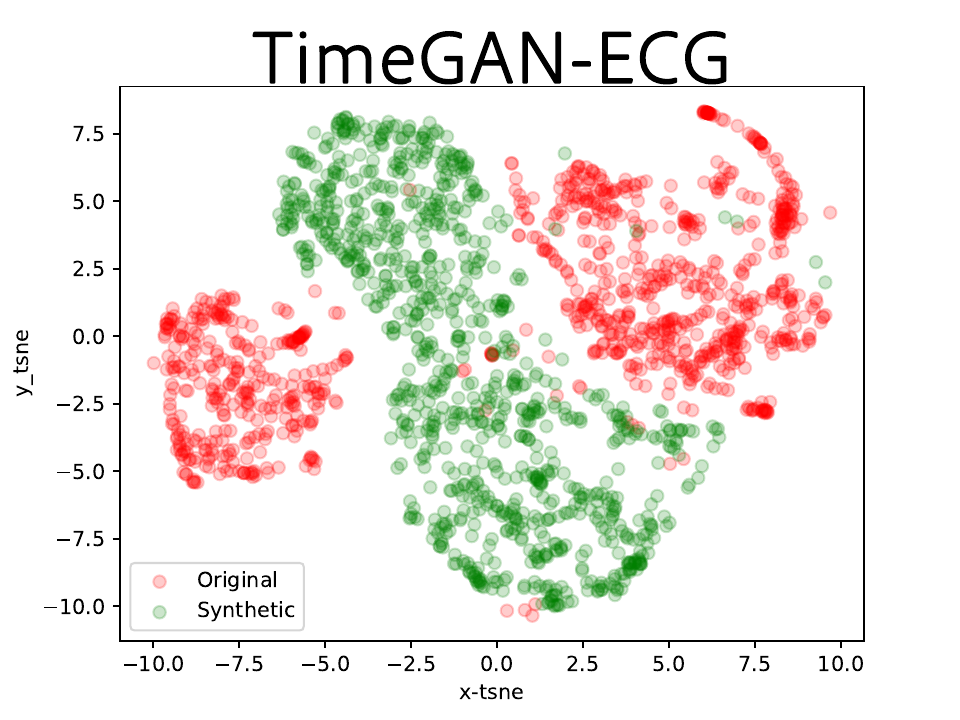}}\hfill
\subfloat{\includegraphics[width=0.122\textwidth]{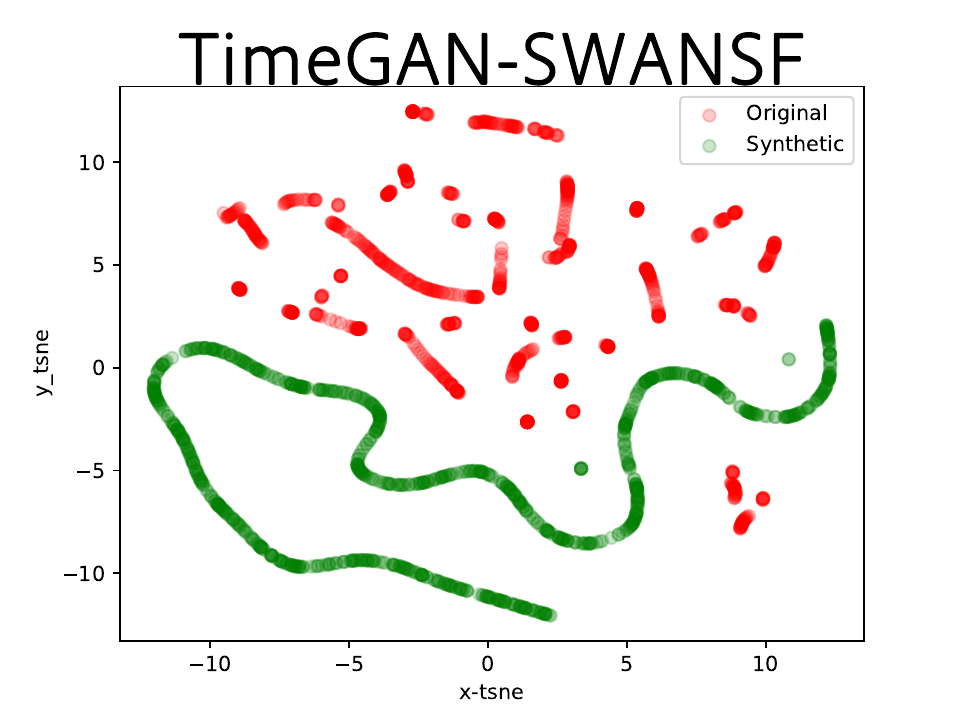}}\hfill

\subfloat{\includegraphics[width=0.122\textwidth]{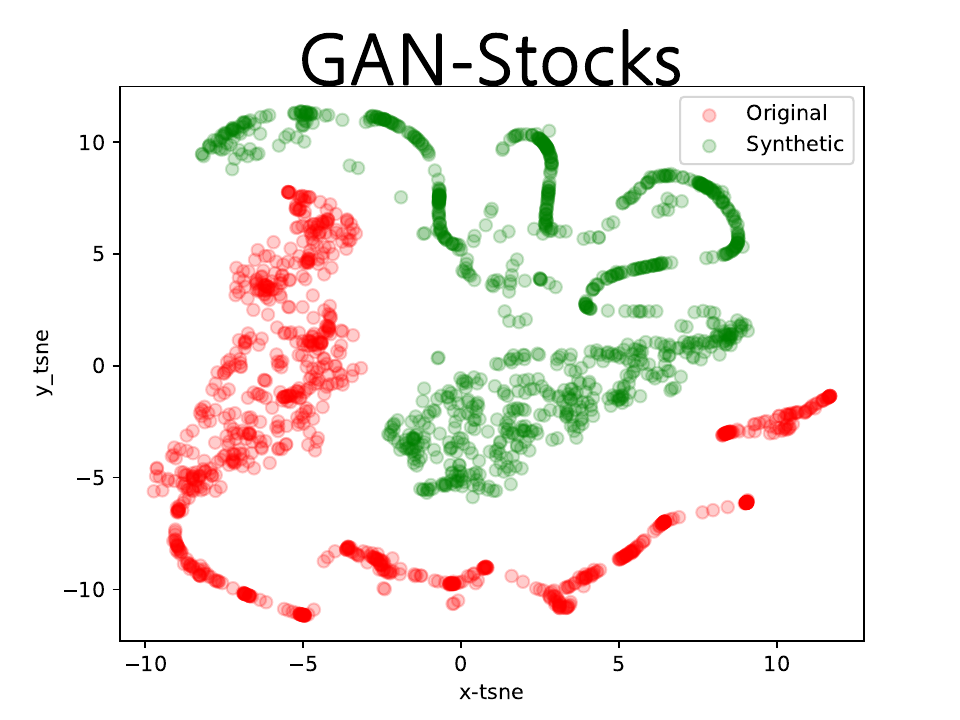}}\hfill
\subfloat{\includegraphics[width=0.122\textwidth]{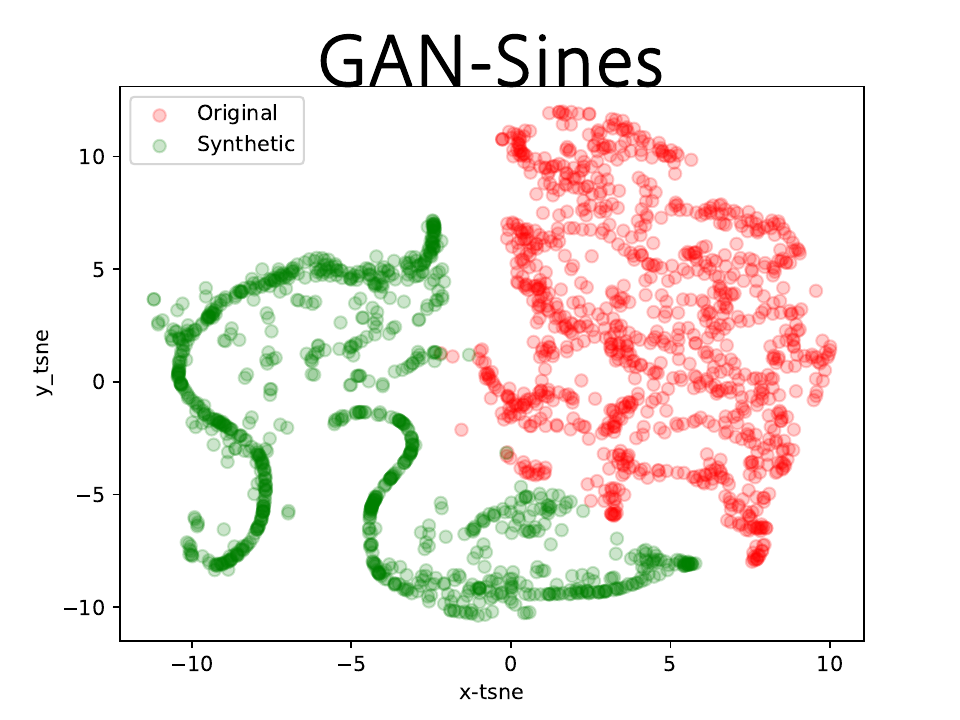}}\hfill
\subfloat{\includegraphics[width=0.122\textwidth]{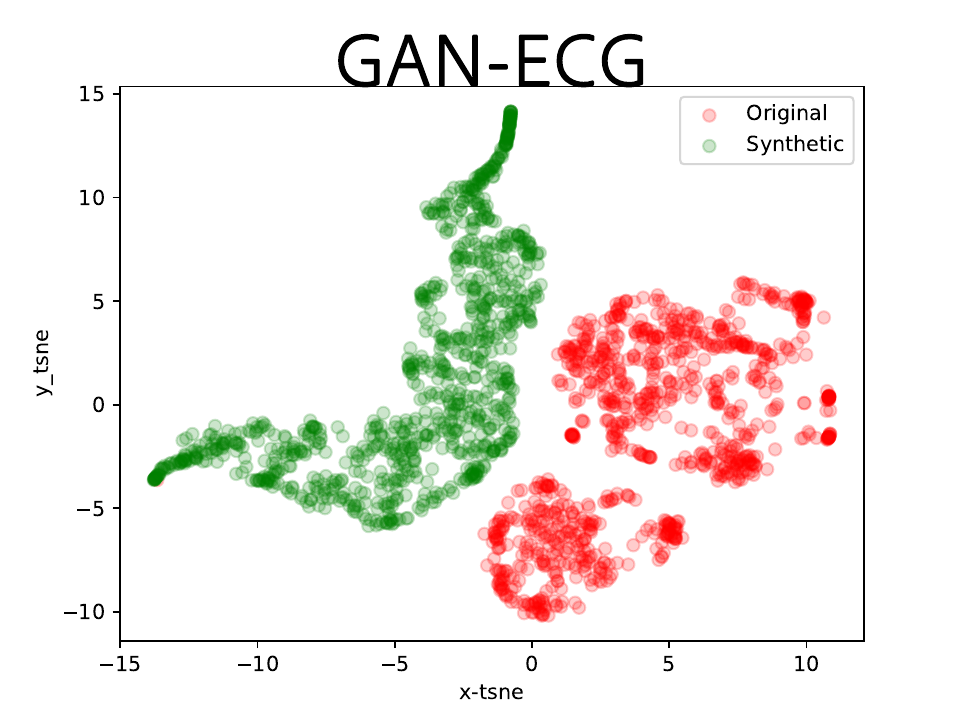}}\hfill
\subfloat{\includegraphics[width=0.122\textwidth]{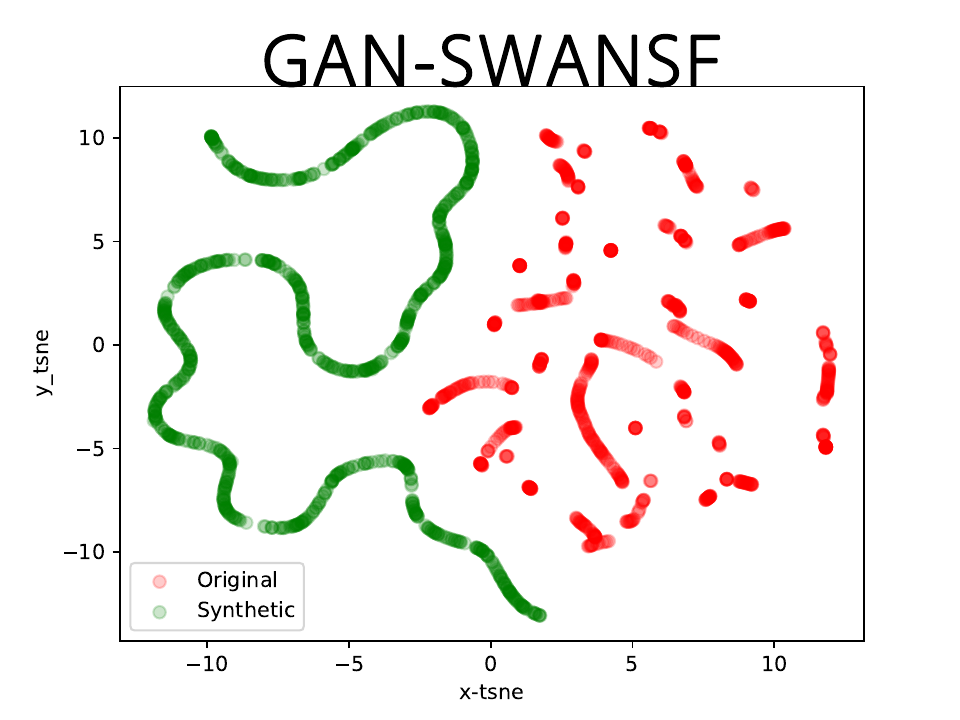}}\hfill

\subfloat{\includegraphics[width=0.122\textwidth]{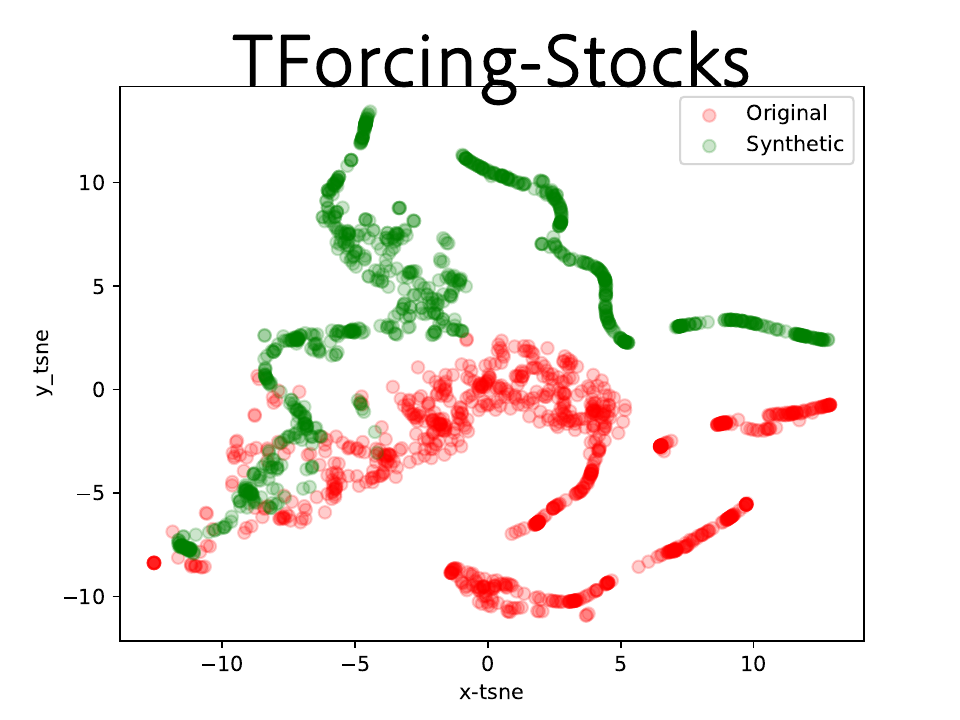}}\hfill
\subfloat{\includegraphics[width=0.122\textwidth]{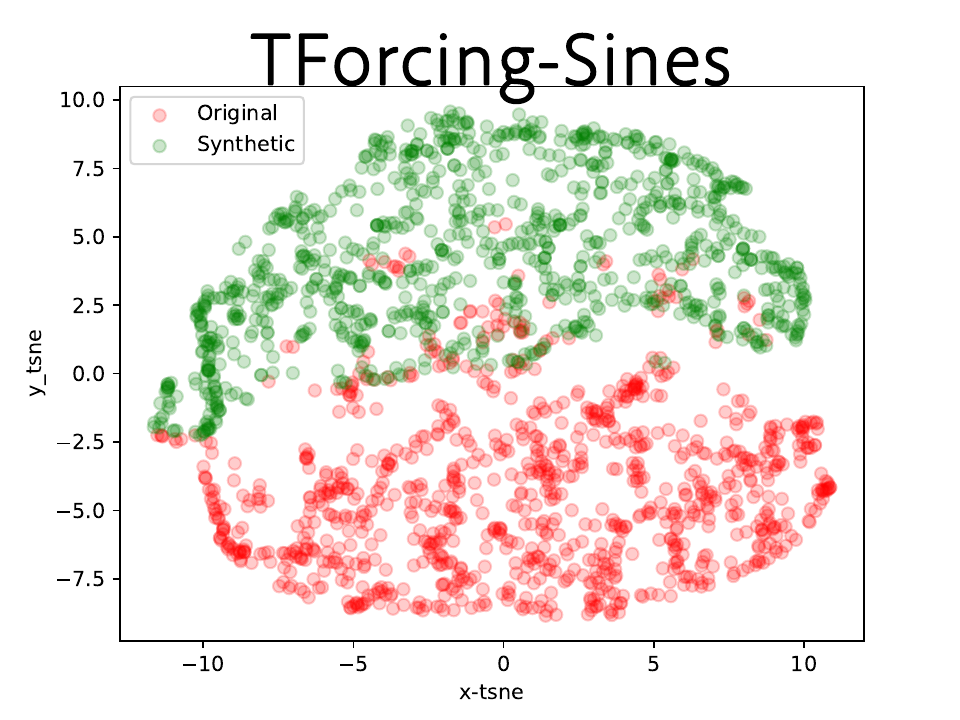}}\hfill
\subfloat{\includegraphics[width=0.122\textwidth]{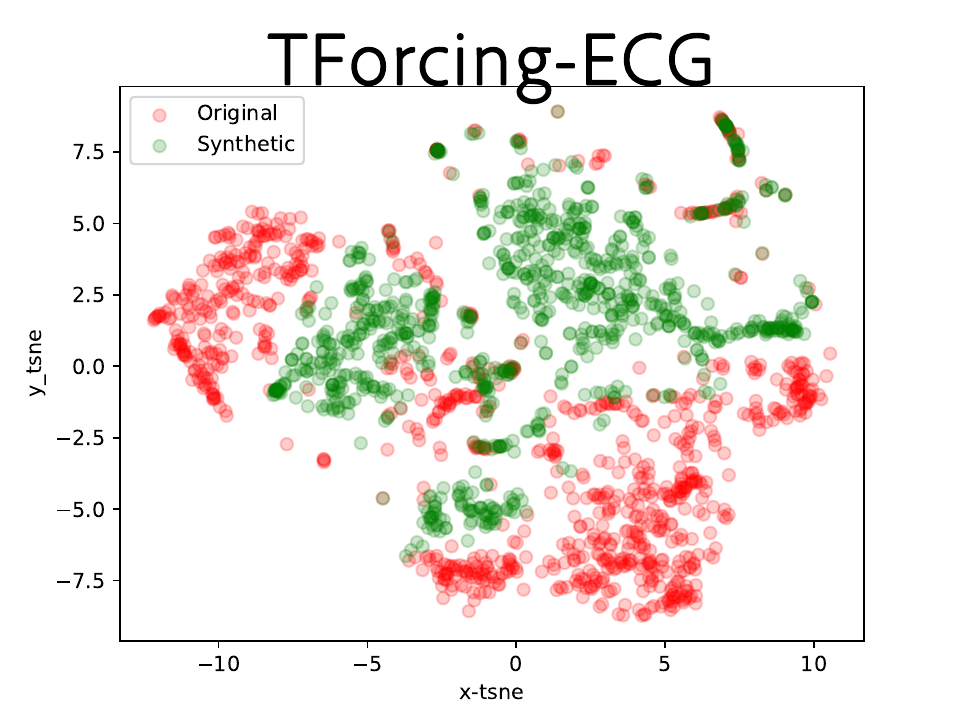}}\hfill
\subfloat{\includegraphics[width=0.122\textwidth]{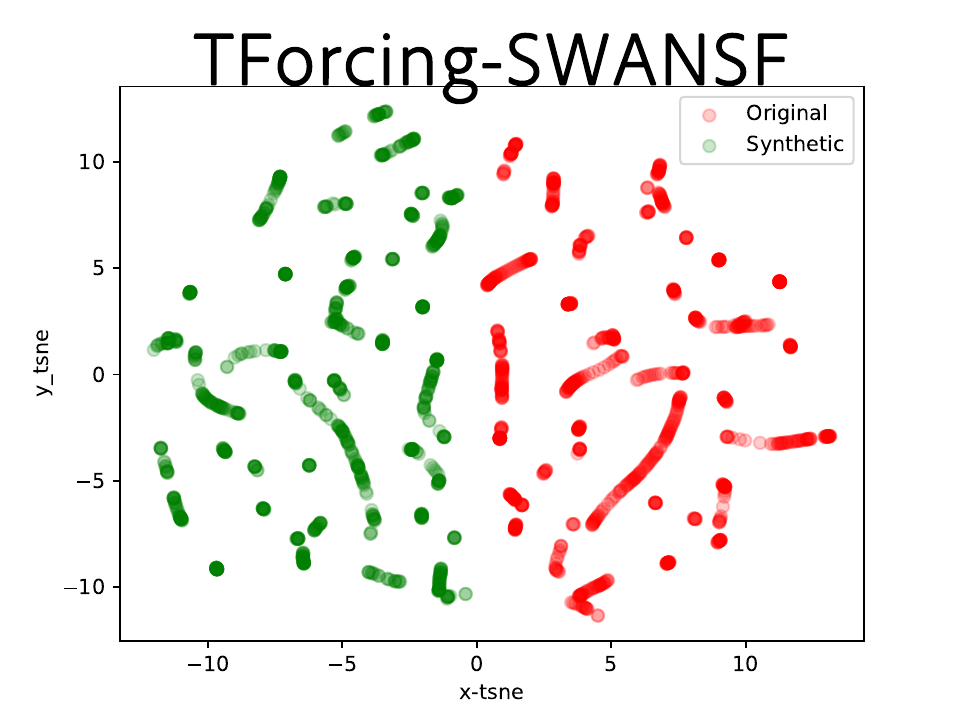}}\hfill

\subfloat{\includegraphics[width=0.122\textwidth]{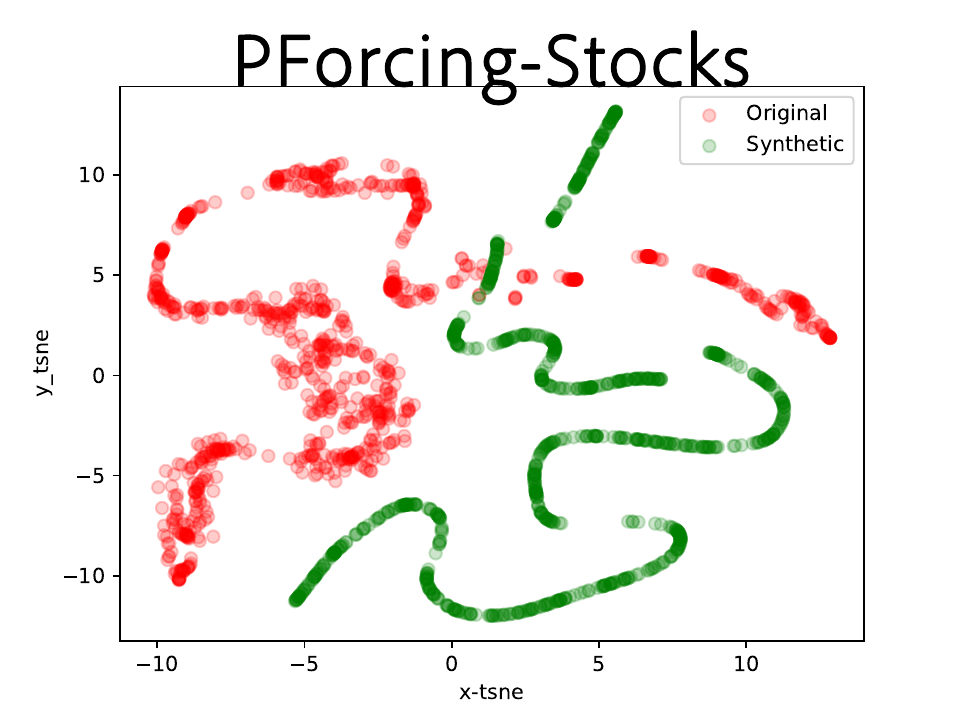}}\hfill
\subfloat{\includegraphics[width=0.122\textwidth]{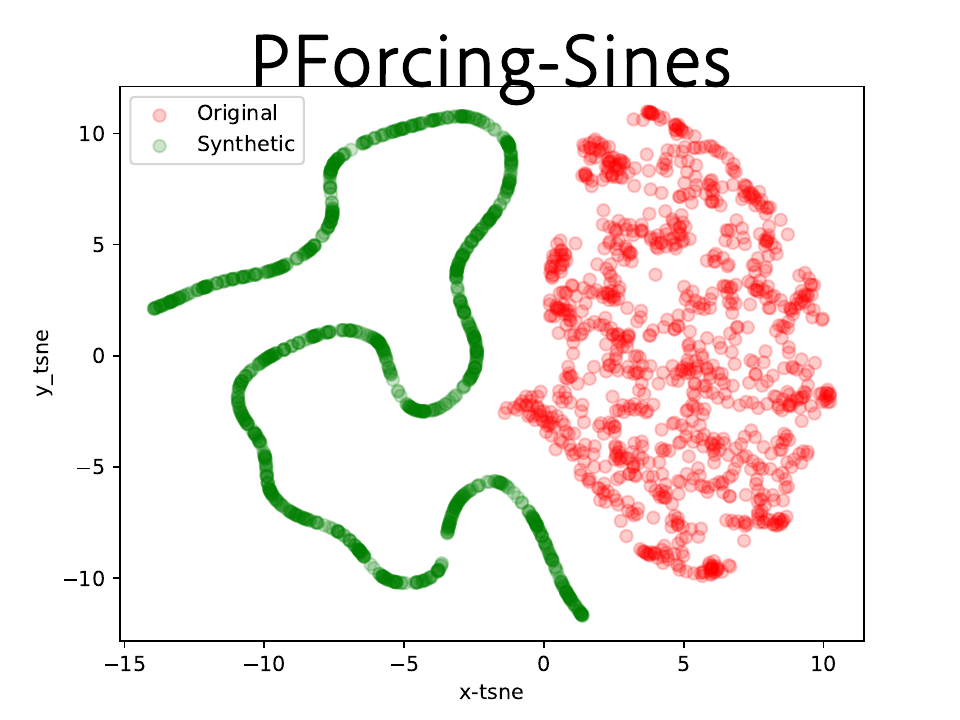}}\hfill
\subfloat{\includegraphics[width=0.122\textwidth]{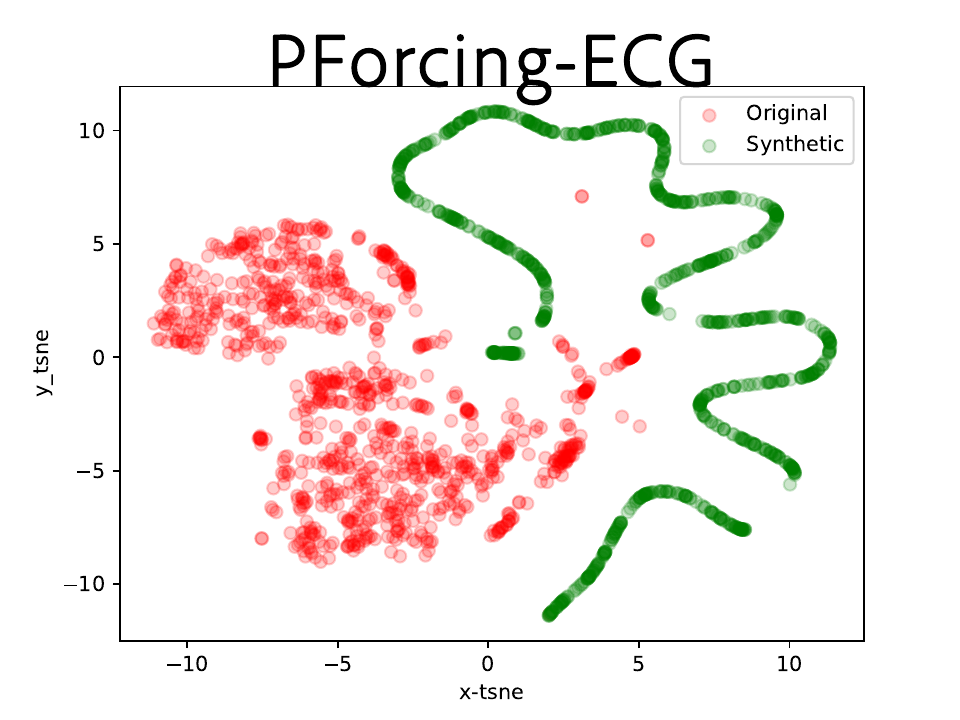}}\hfill
\subfloat{\includegraphics[width=0.122\textwidth]{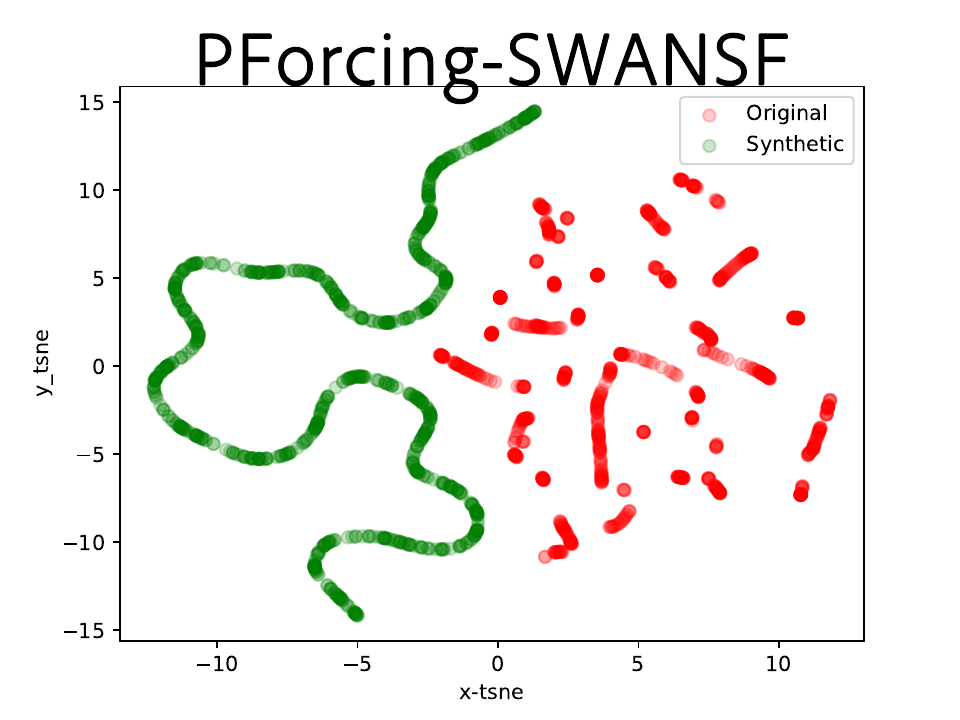}}\hfill

\caption{t-SNE visualizations demonstrate the alignment in distribution between the original and synthetic data samples produced by ChronoGAN and other benchmark models across four datasets.}
\label{fig:sines}
\end{figure}

Based on Figs. \ref{fig:stocks} and \ref{fig:sines}, ChronoGAN demonstrates a superior ability to learn the probability distribution of real datasets more efficiently than all other baseline techniques. This is crucial, as a GAN-based model must generate data that accurately covers the entire distribution of the real dataset. The PCA and t-SNE results for the Stocks dataset show highly accurate outcomes. This achievement is primarily due to the \(\mathcal{L}_V\) loss, which enables the network to effectively capture the mean and variance of each batch of real data.

\balance

\section{Conclusion and Future Work}
In this study, we present ChronoGAN, an innovative model designed for generating time series data. ChronoGAN consists of five networks: an autoencoder (comprising an encoder and decoder), a generator, a supervisor, and a discriminator. These networks are trained together to learn the probability distribution and stepwise temporal dynamics of time series data. The model employs adversarial training in the feature space while generating data in the latent space, which significantly enhances the performance of both the autoencoder and generator networks. Additionally, ChronoGAN introduces novel loss functions for the autoencoder, generator, and supervisor networks, along with a new neural network architecture and an early generation mechanism. This framework consistently outperforms leading methods in generating realistic time series data, both qualitatively and quantitatively. In future research, we aim to integrate these concepts into adversarial autoencoders to develop an advanced framework for producing high-quality time series data.

\section{Acknowledgment}

Support for this work has been provided by the Division of Atmospheric and Geospace Sciences within the Directorate for Geosciences through NSF awards \#2301397, \#2204363, and \#2240022, as well as by the Office of Advanced Cyberinfrastructure within the Directorate for Computer and Information Science and Engineering under NSF award \#2305781.

\end{document}